\pdfoutput=1

\documentclass[11pt]{article}

\usepackage[final]{acl}

\usepackage{times}
\usepackage{comment}
\usepackage[colorinlistoftodos]{todonotes}
\usepackage{latexsym}
\usepackage{booktabs}
\usepackage{graphicx}
\usepackage{hyperref}
\usepackage{graphicx}
\usepackage{multirow}
\usepackage{tikz}
\usetikzlibrary{positioning, shapes, arrows.meta}
\usepackage{pifont}  
\usepackage{xcolor}  
\usepackage{fontawesome5}  
\usepackage[T1]{fontenc}

\usepackage[utf8]{inputenc}

\usepackage{microtype}

\usepackage{inconsolata}
\usepackage{graphicx}
\usepackage{subfig}
\usepackage{graphicx}

\title{Knowing Before Saying: LLM Representations Encode Information About Chain-of-Thought Success Before Completion}

\author{Anum Afzal \\
  Technical University of Munich \\
  \texttt{anum.afzal@tum.de} \\ 
  \And
   Florian Matthes \\
  Technical University of Munich \\
  \texttt{matthes@tum.de} 
  \AND
  Gal Chechik \\
  Nvidia Research \& Bar-Ilan University
 \\
  \texttt{gchechik@nvidia.com} \\ \\\And
  Yftah Ziser \\
  Nvidia Research \\
  \texttt{yziser@nvidia.com} \\}

\begin{document}
\maketitle
\begin{abstract}
We investigate whether the success of a zero-shot Chain-of-Thought (CoT) process can be predicted before completion. We discover that a probing classifier, based on LLM representations, performs well \emph{even before a single token is generated}, suggesting that crucial information about the reasoning process is already present in the initial steps representations. In contrast, a strong BERT-based baseline, which relies solely on the generated tokens, performs worse—likely because it depends on shallow linguistic cues rather than deeper reasoning dynamics.
Surprisingly, using later reasoning steps does not always improve classification. When additional context is unhelpful, earlier representations resemble later ones more, suggesting LLMs encode key information early. This implies reasoning can often stop early without loss.
To test this, we conduct early stopping experiments, showing that truncating CoT reasoning still improves performance over not using CoT at all, though a gap remains compared to full reasoning. However, approaches like supervised learning or reinforcement learning designed to shorten CoT chains could leverage our classifier’s guidance to identify when early stopping is effective. Our findings provide insights that may support such methods, helping to optimize CoT’s efficiency while preserving its benefits.\footnote{Code and data is available at \href{https://github.com/anum94/CoTpred}{\texttt{github.com/anum94/CoTpred}}.}
\end{abstract}

\section{Introduction}
Chain-of-Thought (CoT)  prompting \cite{wei2023chainofthoughtpromptingelicitsreasoning} enhances the capability of large language models (LLMs) to perform multi-step reasoning. It explicitly guides the LLM in creating intermediate explanations to solve a problem, offering a sequence of reasoning steps while responding to a prompt. Given its effectiveness, CoT has found success in mathematical reasoning \cite{zheng-etal-2023-chain}, medical applications \cite{liu-etal-2024-medcot}, faithfulness evaluation \cite{xu-etal-2024-faithful}, and multimodal models \cite{wang-etal-2024-videocot,kumari-etal-2024-m3hop,byun-etal-2024-ares}.
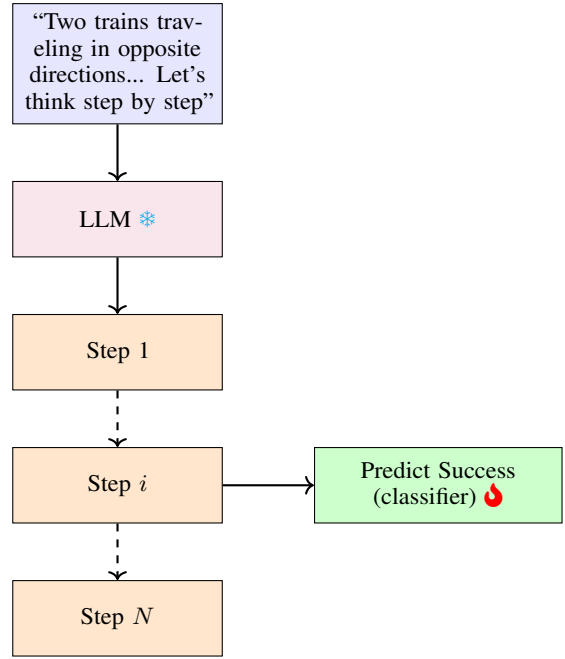
\begin{figure}[t]
    \centering
    \begin{tikzpicture}[
        every node/.style={font=\small, text centered}, 
        box/.style={draw, minimum width=2.5cm, minimum height=1cm, align=center, fill=blue!10, text width=2.5cm},
        llmbox/.style={draw, minimum width=2.5cm, minimum height=1cm, align=center, fill=purple!10, text width=2.5cm},
        stepbox/.style={draw, minimum width=2.5cm, minimum height=1cm, align=center, fill=orange!20, text width=2.5cm},
        predict/.style={draw, minimum width=3.0cm, minimum height=1cm, align=center, fill=green!20, text width=3.0cm},
        arrow/.style={->, thick}
    ]
    
        \node[box] (problem) {``Two trains traveling in opposite directions... Let's think step by step''};
        \node[llmbox, below=0.75cm of problem] (llm) {LLM \textcolor{cyan}{\ding{100}}};
        \node[stepbox, below=0.75cm of llm] (step1) {Step 1};
        \node[stepbox, below=0.75cm of step1] (stepi) {Step \( i \)};
        \node[stepbox, below=0.75cm of stepi] (stepN) {Step \( N \)};
        \node[predict, right=1.2cm of stepi] (predict) {Predict Success\\(classifier) \textcolor{red}{\faFire}};
        \draw[arrow] (problem) -- (llm);
        \draw[arrow] (llm) -- (step1);
        \draw[arrow, dashed] (step1) -- (stepi);
        \draw[arrow, dashed] (stepi) -- (stepN);
        \draw[arrow] (stepi) -- (predict);
    \end{tikzpicture}
    \caption{Illustration of our approach. The LLM generates intermediate reasoning steps in a Chain-of-Thought sequence. At step \( i \), we use its internal representations to predict whether the CoT process will succeed. The snowflake (\textcolor{cyan}{\ding{100}}) indicates frozen parameters, while the flame (\textcolor{red}{\faFire}) indicates trainable parameters.}
    \label{fig:cot_prediction}
\end{figure}
While CoT reasoning has been shown to improve performance across many tasks, it is computationally expensive, as it requires decomposing complex problems into a series of intermediate steps, each demanding its own processing. This raises two intriguing questions: a) Do LLMs implicitly "know" whether they will arrive at a correct answer before completing their reasoning? and b) If progressing past the initial steps doesn't improve this knowledge, does this indicate that the LLM has completed its calculation? Given the high computational cost of CoT, understanding when and how LLMs "know" their answer could enable more efficient and targeted reasoning strategies. Developing a method to assess whether CoT will lead to a correct conclusion could optimize resource allocation—stopping reasoning early when the outcome is clear or dedicating more steps when uncertainty remains. Furthermore, this knowledge could inform annotation efforts to support CoT-specific fine-tuning. 

To explore these questions, we create a CoT success prediction dataset derived from popular math-solving datasets, where zero-shot CoT has been shown to significantly outperform vanilla prompting (more details are provided in Section \ref{section:datasets}). By applying a CoT prompt to these questions, we capture the LLM activations across multiple CoT steps and annotate the final answer as either correct or incorrect. We then train a probing classifier \cite{belinkov2022probing} on top of the LLM representations (assuming white-box access) to predict if a given prefix of the ongoing CoT sequence would lead to a correct answer (see Figure \ref{fig:cot_prediction}). Our experiments show that by leveraging the LLM's internal representations, our classifier can effectively predict whether a CoT sequence will be successful—even before generating a single token—achieving 60\% to 76.4\% accuracy across different datasets and LLMs. Notably, it outperforms BERT \cite{devlin-etal-2019-bert}, a strong text classifier that relies only on the input text, demonstrating that the LLM’s internal representations encode valuable information about intermediate calculations—information that BERT, constrained to shallow linguistic features, cannot capture. 

Further experiments with mid-CoT steps reveal an intriguing pattern: in two of six cases, providing the classifier with later reasoning steps does not significantly improve its prediction accuracy. Using SVCCA, a complementary method to probing, we find that in these scenarios, earlier steps are more similar to the final step compared to the dataset where additional CoT context benefits the classifier. We conduct initial zero-shot experiments to investigate whether this similarity allows for early termination of the CoT process without affecting the final answer. While zero-shot alone is insufficient, our results suggest that more targeted approaches, such as supervised learning or reinforcement learning, could effectively shorten the CoT process while maintaining strong performance.

To conclude, our contributions are as follows:
\begin{enumerate}
\item We define the task of Chain-of-Thought (CoT) success prediction and investigate whether LLMs inherently estimate the effectiveness of CoT prompting before generating a full answer.
\item We construct datasets for this task and train a lightweight probe that predicts the success of CoT prompting before the LLM completes its generation.
\item Through extensive analysis, we demonstrate that leveraging LLM internal representations significantly improves classification performance compared to relying solely on generated tokens, indicating that these representations capture knowledge about intermediate calculations.
\item We conduct initial experiments on zero-shot early stopping in CoT, showing that while a gap remains between early stopping and full CoT completion, stopping mid-calculation still slightly outperforms not using CoT at all. This suggests that stronger methods could further unlock LLMs' potential to shorten CoT chains while maintaining high performance.
\end{enumerate}

\section{Related Work}
We review studies that analyze and customize CoT reasoning and those that use the internal LLM representations to predict aspects of their generation in advance. Given the breadth of research in these areas, we focus on the most relevant works for our setup.

\subsection{Analyzing Chain-of-Thought}
As CoT reasoning gained popularity, an increasing number of papers sought to analyze its underlying mechanisms, mainly when applied to solving math questions, where it often excels. \citet{xu-etal-2024-numcot} demonstrated how minor changes
in numbers or units can drastically affect CoT performance. Several empirical studies have explored key factors in improving CoT performance. For example, \citet{madaan2022text} used counterfactual prompts to highlight the importance of symbolic reasoning in the CoT process. \citet{wang-etal-2023-towards} found that the order of rationales and their relevance to the query are the most crucial aspects of CoT. \citet{rai-yao-2024-investigation} analyzed the neurons in LLM feed-forward layers to determine whether information about the design decisions studied in empirical research is encoded within them. Alternatively, \citet{liu2024mindstepbystep} draw a comparison between an LLM  and a human's overthinking nature. They show a series of tasks where similar to humans, LLM's performance also gets worse with CoT prompting. 

\citet{wang-etal-2023-towards} evaluate which factors play a role in CoT prompting and show that even using incorrect demonstration in CoT prompting leads to the generation of correct answers. In contrast, \citet{cui2024theoreticalunderstandingchainofthoughtcoherent} show that errors in intermediate reasoning steps tend to affect CoT performance. Lastly, \citet{pfau2024let} used filler words like "..." to show that the CoTs displayed by LLM are just superficial and rather similar internal compute responsible for LLM's reasoning could be triggered by meaningless filler words. \citet{bao-etal-2025-likely} use structural causal models to analyze how instructions, reasoning steps, and answers interact in CoT prompting. Instead, our work probes internal representations to assess whether models encode information about answer correctness during the reasoning process. While their analysis is causal and text-level, ours focuses on what the model "knows" internally.

\subsection{Probing LLMs Internal Representations}
The internal representations of LLM have been used to gather insights about tasks; one such task is using representations at a given state $t$ to predict the words at positions beyond $t + 2$ \cite{goyal2024thinkspeaktraininglanguage}. \citet{NEURIPS2023_ed3fea90} try to evaluate if LLM are honest in their explanations and concluded that LLM explanations are heavily biased by simple variations in the prompt, such as reordering of items. \citealp{azaria-mitchell-2023-internal,gottesman-geva-2024-estimating,seodetecting} investigate the task of estimating LLM's knowledge on a given subject before it starts generation. They approach this task by using the internal representations of LLM as training features for a probe that can predict if the LLM output would be faithful. The field of probing to comprehend the internal mechanisms of CoT using LLM's internal representation is still evolving. 

\subsection{CoT Reasoning for Math Tasks}
Recent research has demonstrated the effectiveness of CoT reasoning in logical tasks \cite{sprague2024cotcotchainofthoughthelps}, particularly for solving mathematical datasets. \citet{ahn-etal-2024-large} examine the latest advancements in large language models (LLMs) for mathematical reasoning and highlight the effectiveness of CoT in this domain. Furthermore, \citet{ji2025mygomultiplexcotmethod} introduced a dual CoT approach that incorporates self-reasoning and self-criticism to improve performance on math tasks.
Several other studies \cite{wang2023mathcoderseamlesscodeintegration,li2023camel} have focused on improving LLM performance in mathematical tasks by developing custom models optimized for benchmark datasets.

\section{Methodology}
\paragraph{Constructing the Dataset}
We first generate deterministic outputs to assess whether an LLM inherently "knows" if it can solve a task using CoT prompting. We run inferences with a temperature of zero, ensuring consistency and eliminating stochastic noise. In addition, low sampling temperatures are recommended for tasks that require precision and factual accuracy, such as technical writing, code generation, or question answering, which is particularly crucial for solving math problems \cite{renze-2024-effect}. Each generated response is compared to a reference answer, assigning a correctness label that serves as a ground-truth label for our trained classifier. Next, assuming white-box access to the LLM, we extract the LLM’s hidden states from the initial forward pass of the prompt to examine whether its internal representations encode predictive information about CoT success. These hidden states ($H$) capture the LLM's pre-generation reasoning and are used as training features. For each sample, we obtain a 3D tensor ($H$, $N$, $k$), where $H = {h_1, h_2, \dots, h_L}$ represents the hidden layers, $N$ is the number of samples, and $k$ is the hidden dimension size. Since prompt lengths vary, we use the last token’s representation for consistency. To study the contribution of different layers, we train a separate probe for each hidden layer, evaluating its accuracy as a measure of the layer’s role in CoT success prediction. A higher classification accuracy ($C_L$) indicates that layer $L$ contains more information about the likelihood of success. We conduct our experiments using \texttt{Llama-3.1-8B} and \texttt{Mistral-7B}, utilizing all hidden layers, each with 4096 dimensions.

\paragraph{Classification Model}Following the precedent set by \citet{azaria-mitchell-2023-internal}, we employ a compact feedforward neural network with three hidden layers of 256, 128, and 64 units, each using ReLU activation. The output layer applies a sigmoid activation function. Training is optimized using either the Adam or SGD optimizer, which is selected based on empirical performance for each dataset. We perform a hyperparameter tuning specific to the task and detail the selected configurations in \autoref{app:technical-details}. The classifier is trained for five epochs across all datasets and LLMs.

\paragraph{Success Prediction over Time}
While our primary methodology focuses on predicting CoT performance before LLM generation begins, we also explore how this prediction may change during the generation process. Identifying atomic steps in the generation can be challenging due to their varying lengths and styles, so for simplicity, we instead use percentages of the total number of tokens generated (10\%, 20\%, up to 90\%) and concatenate them with the initial prompt (see \autoref{app:dataset-examples} for illustration). This approach allows for a consistent evaluation across different generations. As in the previous approach, we extract hidden layers for partial generations at 10\% intervals, enabling us to observe the evolution of internal representations. We apply a similar strategy to expand our test sets.

\begin{table}[!htpb]
    \centering
     \resizebox{\columnwidth}{!}{
    \begin{tabular}{ccccc}
        
        \textbf{CoT} & \textbf{wo CoT} &\multicolumn{3}{c}{\textbf{success rate}}\\
           \midrule
   & & \textbf{Cn-k12} &\textbf{AQuA} & \textbf{Olympiad} \\
           \toprule      
    \multicolumn{5}{c}{\textit{\texttt{Llama-3.1-8B}}} \\
        \toprule
        0 & 0 & 41.83\% & 47.46\% & 44.76\% \\
         0 & 1 & 8.11\% & 2.53\%  & 4.77\%\\
        1 & 0& 32.52\% & 43.32\% & 36.97\%\\
        1& 1 & 17.44\% & 6.67\% & 12.92\% \\
        
        \toprule      
    \multicolumn{5}{c}{\textit{\texttt{Mistral-7B}}} \\
        \toprule
        0 & 0 & {44.07\%} & {39.43\%} & {41.43\%} \\
         0 & 1 & {5.89\%} & {10.54\%} & {8.46\%}\\
        1 & 0& {32.33\%} & {33.83\%} & {28.08\%}\\
        1& 1 & {17.69\%} & {16.2\%} & {22.03\%}\\
        \bottomrule
    \end{tabular}
    }
\caption{We depict the percentage of Problems the LLM was able to solve with and without any Chain-of-Thought prompting on a balanced dataset where CoT prompting helped solve 50\% of the problems. We show a confusion matrix such that 0 means LLM was not able to solve the problem, and 1 means that it was able to solve the task. }

\label{tab:cot-vs-no-cot}
\end{table}


\section{Experimental Setup}

\subsection{Baseline}
The prediction of CoT success may depend more on linguistic cues in the text than on internal LLM representations encoding intermediate computations. To explore this, we build on \cite{azaria-mitchell-2023-internal}, which examined whether LLMs store information in their internal states, albeit for different purposes. Our goal is to demonstrate that such information is indeed retained within the LLM's representations. To test this, we use BERT as a baseline, as it relies solely on textual tokens without access to internal LLM states, effectively functioning as a black-box access approach. Given BERT's strength as a text classifier, its ability to predict CoT success would indicate that textual cues alone suffice. We use the \texttt{google-bert/bert-base-uncased} variant with default settings, maintaining a consistent neural network-based classification setup while varying the input features.\footnote{Training features vary when using BERT embeddings (dimension 768) or LLM layers (dimension 4096).}
\subsection{Large Language Models}
For the hidden representations of \texttt{Llama-3.1-8B} and \texttt{Mistral-7B}, we use the \texttt{meta-llama/Llama-3.1-8B-Instruct} and \texttt{mistralai/Mistral-7B-Instruct-v0.3} checkpoints on huggingface respectively. 
\subsection{Datasets}
\label{section:datasets}
We used three math problem datasets of varying difficulty in our experiments: World Olympiads Data (Olympiad) \cite{numina_math_datasets}, Chinese K-12 Exam (cn-k12) \cite{numina_math_datasets}, and AQuA (aqua) \cite{ling2017program}. To assess model behavior across different reasoning patterns, we ran each dataset with two different LLMs—\texttt{Llama-3.1-8B} \cite{grattafiori2024llama3herdmodels} and \texttt{Mistral-7B}—resulting in six distinct dataset variants. Since we enforce class balance (i.e., an equal number of correct and incorrect generations), the two versions of the same original dataset may contain different sets of questions, depending on the LLM's outputs. Given the difficulty level of questions within each dataset, the success rate using \texttt{Llama-3.1-8B}, for example, varies considerably, reaching 22\%, 28\%, and 62.3\% on Olympiad, cn-k12, and AQuA, respectively. For all datasets and both LLMs, we ran inference to obtain balanced train (10k) and test (1k) sets with an equal distribution of positive and negative examples. Consequently, our classification model is trained on a $10000 \times 4096$ feature space for each dataset and LLM layer combination. We reserve 10\% of the training set for validation to tune hyperparameters. The distribution of question and generation lengths is summarized in \autoref{tab:ds_dist}. Lastly, it is also important to evaluate how well the same LLM performs when solving the same problems without chain-of-thought (CoT) prompting. We present this comparison in a confusion matrix in \autoref{tab:cot-vs-no-cot}, which shows that, for instance, without CoT, \texttt{Llama-3.1-8B} achieved only 9.2\%, 17.69\%, and 25.55\% success rates on AQuA, Olympiad, and cn-k12, respectively, substantially lower than the 50\% CoT accuracy we enforce when collecting a balanced dataset.

\begin{table}
\small
    \centering
     \resizebox{\columnwidth}{!}{
    \begin{tabular}{lrrrrrr}
       
         & \multicolumn{3}{c}{\textbf{\# Question Tokens}} & \multicolumn{3}{c}{\textbf{\# Generation Tokens}}\\
          \cmidrule(lr){2-4} \cmidrule(lr){5-7}
        Dataset & avg & min & max & avg & min & max \\
        \toprule      
    \multicolumn{7}{c}{\textit{\texttt{Llama-3.1-8B}}} \\
        \toprule
        AQuA      & 42.96 & 5  & 308 & 287.24 & 17 & 512\\
        Olympiad   & 67.76 & 3 & 1109 & 401.56 & 2 & 512 \\
        Cn-k12  & 76.38 & 7  & 520 & 327.46& 4 & 512 \\

        \toprule
            \multicolumn{7}{c}{\textit{\texttt{Mistral-7B}}} \\
        \toprule
        AQuA      & 84.23 & 38  & 344 & 255.07 & 35 & 509\\
        Olympiad   & 86.10 & 9 & 1108 & 368.47 & 40 & 511 \\
        Cn-k12  & 88.94 & 5  & 679 & 304.96& 29 & 511 \\

        \bottomrule
    \end{tabular}
    }
\caption{Average, Minimum and Maximum token count of Questions and LLM generation using \texttt{Llama-3.1-8B} and \texttt{Mistral-7B} tokenizers on all three datasets. Generation Tokens are capped at 512 using max\_new\_tokens = 512.}
\label{tab:ds_dist}
\end{table}

\begin{table*}[!htbp]

 \centering
 \resizebox{\textwidth}{!}{
 \begin{tabular}{lcccccccccccc}
     &  \multicolumn{2}{c}{\textbf{Before Gen}} & \multicolumn{9}{c}{\textbf{Over Time}}\\
     \cmidrule(lr){2-3}\cmidrule(lr){4-13}\\
    Dataset  & top-5 layers & 0\%  &  10\% & 20\% & 30\% & 40\% & 50\% & 60\% & 70\%& 80\% & 90\% & 100\%\\
           \toprule      
    \multicolumn{13}{c}{\textbf{\textit{\texttt{Llama-3.1-8B} (human-annotated)}}} \\
        \toprule
   \multicolumn{13}{c}{\textit{BERT (baseline)}} \\
   \midrule
   AQuA & - & {53.50} & 54.2 & {55.8} & {54.1} &{54.1} & {53.5} & {51.7} &{51.6}& {53.7} &{51.7} & {50.6}\\
   Olympiad & - &  {69.10} & {67.9} & {68.7}& {66.2} &{66.4} & {67.4} & {65.1} &{67.9}& {66.8} &{67.1} & {66.9}\\
   Cn-k12 & - & {66.20} & {57.8}& {64.4} & 64.1 & 63.8 &64.3&63.8 & 62.7& 61.7& {61.1} &{62.4}\\
   \bottomrule
    \multicolumn{13}{c}{\textit{Our Model}} \\
   \midrule
   AQuA & {11, 12, 13, \textbf{14}, \textbf{16}} &{60}  & {51.5} & {60} & {60.5}& {63.2}& {60.11}& {60.8} & {63.7} &{62.9} &{65.7}&  \textbf{69.4}\\
   Olympiad & {8, \textbf{14}, \textbf{16}, 17, 31} & {76.4} & {75.1}&  {75.8} & {73.8}  &{74.6}& \textbf{76.5}&  {73.8}& {75.3} &{75.3} &{73.8} &{75.9} \\
   Cn-k12 & {13, \textbf{14}, \textbf{16}, 17 22} & {69.10} & {67.7}&{69.9}& {69.2} &{67.7} & {67.2}& {70} & \textbf{71.6}& {70.7} &{68.9} & {70.9}\\
    \toprule      
    \multicolumn{13}{c}{\textbf{\textit{\texttt{Mistral-7B} (\texttt{GPT-4o mini} annotated)}}} \\
        \toprule
    \multicolumn{13}{c}{\textit{BERT (baseline)}} \\
   \midrule
   AQuA & - & {60.1} & 57.8 & {59.5} & {63.4} &{65.2} & {63.7} & {67.8} &{66.7}& {66.8} &{68.1} & {71.8}\\
   Olympiad & - &  {68.8} & {68.1} & {69.6} & {68.8} &{68.3} & {68.6}& {66.4}& {67.1} & {68.3} & {68.6}&{69.6}\\
   Cn-k12 & - & {65.5} & {63.5}& {65.8} & 65.1 & 64.4 &64.8&66.5 & 66.7& 67.3& {67.8} &{68.5}\\
   \bottomrule
    \multicolumn{13}{c}{\textit{Our Model}} \\
   \midrule
   AQuA & {15, 16, \textbf{18}, 23, \textbf{28}} &{64.7}  & {54.4} & {66.1} & {66.9}& {66.4}& {65.1}& {65.8} & {64.1} &{66.6} &{67.4}&  \textbf{80.2}\\
   Olympiad & {7, 9, \textbf{18}, 26, \textbf{28} } & {71.8} & {71.7} & {72.0}& {72.3}& {74.1}& {74.2}& {75.6}& {74.5}& {75.3}& {75.5} & \textbf{75.9} \\
   Cn-k12 & {12, 14, \textbf{18}, 21, 24} & {67.1} & {68.0}&{68.4}& {66.7} &{67.7} & {67.8}& {67.1} & {68.1}& 68.6 &{67.6} & \textbf{71.4}\\
   \bottomrule
 \end{tabular}
 }
\caption{Classification model accuracy before generation begins and as it progresses, measured at different completion percentages. The best-performing layers and time steps are highlighted in bold.}
   \label{tab:pred-scores}
\end{table*}

\paragraph{Olympiad:} The Math Olympiad is a competitive examination designed to evaluate students' mathematical skills and competencies. We use the olympiad dataset from the NuminaMath-CoT \cite{numina_math_datasets} collection of the dataset. This dataset is a collection of problems and respective answers following a CoT format collected from international/national contests as well as forums, books, and summer school materials.

\paragraph{Cn-k12: } This large-scale Chinese K-12 education math exercise dataset was translated and realigned to English using GPT-4. This dataset is also part of the NuminaMath-CoT \cite{numina_math_datasets} collection of datasets, and reference answers follow a CoT format.

\paragraph{AQuA:} This dataset \cite{ling2017program} consists of algebraic math problems, each presented with a step-by-step reference solution. Unlike the other two datasets, it includes multiple-choice options. We use the original dataset released by the authors.

\subsection{Manual Annotation}
Given the large dataset size, we used \texttt{GPT-4o mini}\footnote{\url{https://platform.openai.com/docs/models\#gpt-4o-mini}} to label the training and validation sets for the first language model (\texttt{Llama-3.1-8B}). To ensure the reliability of our evaluation, we manually annotated the test set using annotators selected from a local university’s STEM Master’s program, compensated at 16 euros per hour. Annotators evaluated the correctness of each generation based on the question and a reference answer. On a test set of 1,000 samples, human annotations agreed with LLM-generated ones on 90.9\%, 94.8\%, and 93.4\% for AQuA, Cn-K12, and Olympiad, respectively. These numbers indicate that although LLM-generated labels are generally reliable, the disagreement with human judgment suggests caution when using them for evaluation. For the second language model (\texttt{Mistral-7B}), all datasets (training, validation, and test) were annotated using \texttt{GPT-4o mini}. Given the imperfect agreement observed between human and \texttt{GPT-4o mini} annotations for the first model, these results should be interpreted with appropriate caution.

\section{Results and Discussion}
Since the test set for \texttt{Llama-3.1-8B} was manually annotated by human experts, it is considered more reliable than the \texttt{GPT-4o mini}-annotated test set used for \texttt{Mistral-7B}. As a result, in certain analyses where only one LLM is presented, we focus solely on \texttt{Llama-3.1-8B} to ensure more reliable evaluation.

\begin{figure}[!htbp]
\centering
\makebox[\columnwidth]{ 
  \includegraphics[width=\columnwidth]{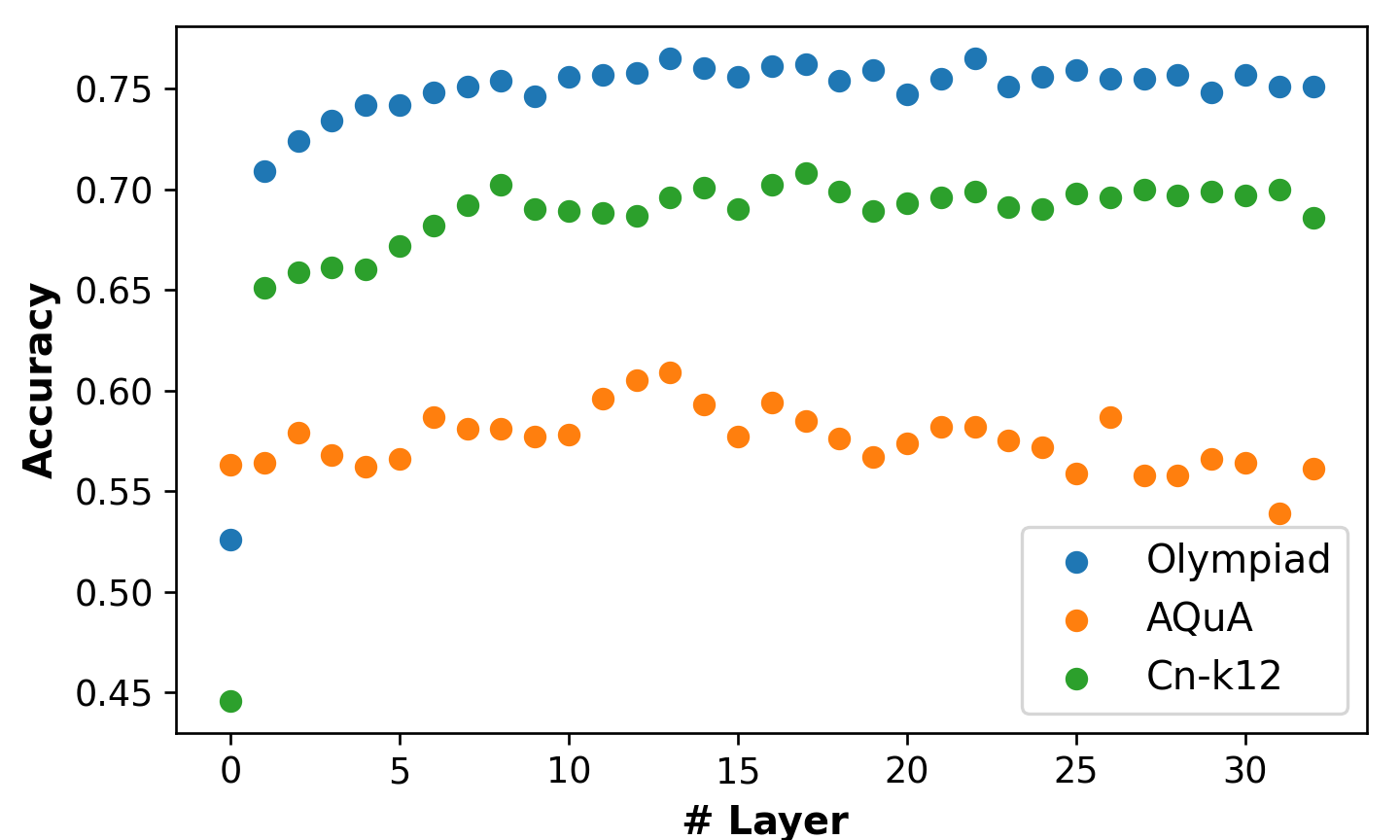}
}
  \caption{Accuracy on the test set when the probe is trained and tested on hidden representations from each layer. Results are shown for all 33 layers of Llama 3.1 8B Instruct across all three datasets.}
  \label{fig:accuracy-per-layer}
\end{figure}

\subsection{Prediction before Generation}
\paragraph{Main Results}
We evaluate the information contained in the internal representation of the LLM before it begins generating and present the results in \autoref{tab:pred-scores}, specifically under the \%0 column. Given that 50\% of answers in the dataset are correct (balanced data), a random classifier would achieve an accuracy of 50\%, which serves as the baseline for interpreting the results. The BERT baseline, which relies exclusively on token representations on CoT prompt with question, consistently outperformed random chance across all six datasets. On generations from \texttt{Llama-3.1-8B}, BERT achieved accuracy scores ranging from 53.5\% to 69.1\%. For \texttt{Mistral-7B}, BERT's accuracy ranged from 60.1\% to 68.8\%. However, the predictive power appears relatively weaker for the AQuA dataset\footnote{We hypothesize that accuracy on \texttt{Llama-3.1-8B}-generated AQuA dataset might be lower as the sampled questions might contain less linguistic patterns than the other datasets.}. Our suggested method, which uses the LLM's internal representations, outperformed the BERT baseline across all datasets, demonstrating their importance for predicting CoT success before generation begins. For instance, on \texttt{Llama-3.1-8B} datasets, our model achieved 60.0\% accuracy on AQuA, significantly higher than BERT, and 76.4\% on Olympiad, compared to BERT's 69.1\%. We observe similar patterns for the \texttt{Mistral-7B} datasets.  The large variance in results (ranging from 60.0\% to 76.4\%) suggests that the effectiveness of the internal representations may be influenced by factors such as task complexity or dataset characteristics, which warrant further investigation. See Appendix \ref{app:confusion-matrix} for details on Classification Performance Breakdown.

\begin{figure*}
  \centering
  \includegraphics[width=0.65\textwidth]{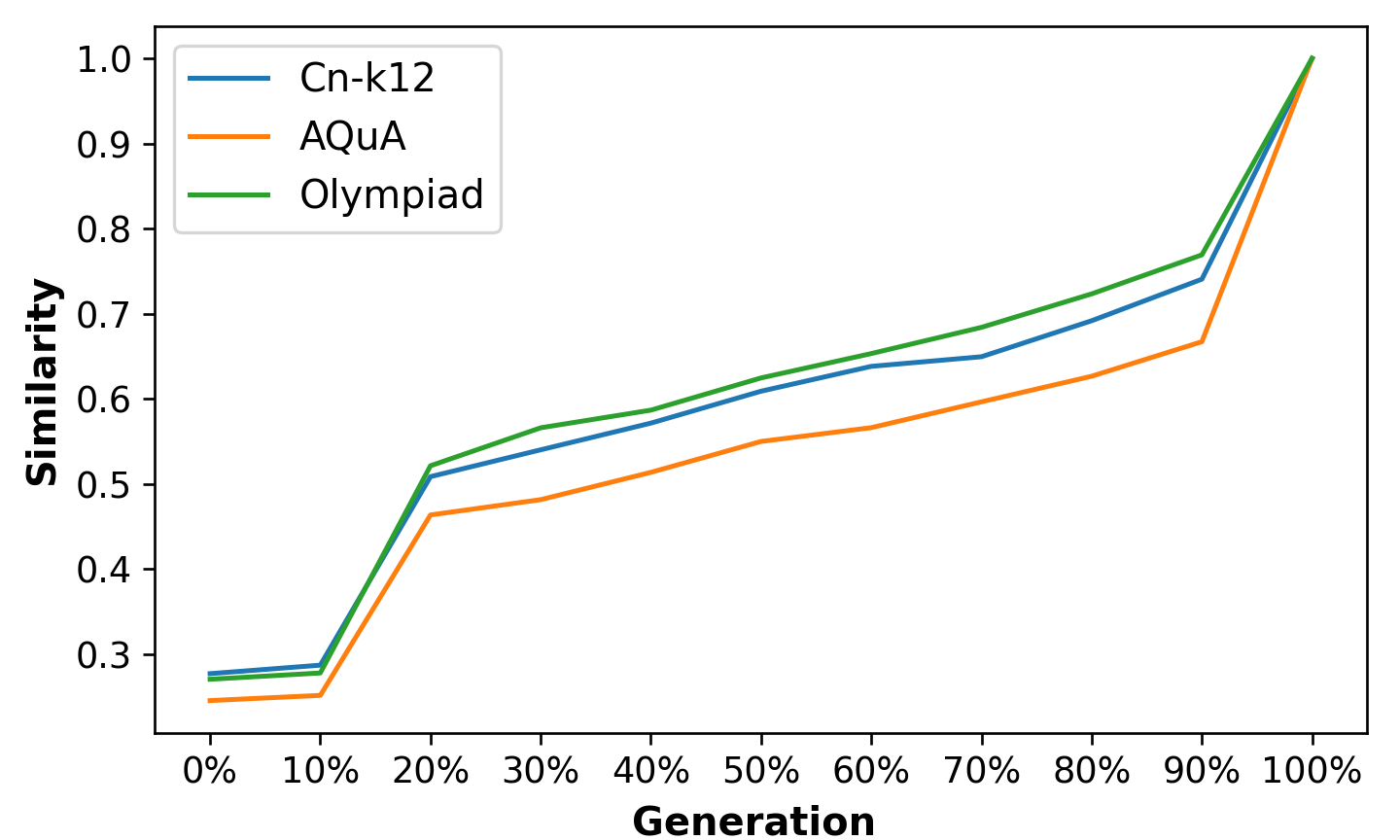}
  
  \caption{Similarity scores using SVCCA between hidden representations of each time step and the full generation, using Layer 14 for all three \texttt{Llama-3.1-8B} datasets.  }
  \label{fig:pot_layer14_graph}
\end{figure*}
\paragraph{LLM Layers Analysis}

In our experiments, we evaluated the accuracy of the classification model when trained on the hidden representations of each layer. We show the accuracy per layer of \texttt{Llama-3.1-8B} in \autoref{fig:accuracy-per-layer}. It can be seen that the middle such as layers 11 - 14 and layers 16 - 17, and in some cases the last layers of the LLM seem to play a role in it's internal notion of Success or Failure. As the in shown \autoref{tab:pred-scores}, we find layer 14 and layer 16 to be consistent among all three datasets suggesting these layers to be more involved in the notion of CoT prediction. Our findings are inline with those of \citealt{azaria-mitchell-2023-internal}, who show similar results regarding LLM's notion of truthfulness and show layer 16 and in some cases the last layers of LLM to be most knowledgable for their task.

\subsection{Prediction over Time}
\paragraph{Main Results}
We evaluate the information contained in the internal representation of the LLM during generation, as shown in \autoref{tab:pred-scores}. To do this, we capture the hidden states at various time intervals—after the LLM has generated 10\%, 20\%, and so on of the total content—and use these states to train the probe. The goal is to assess whether the LLM's understanding of success or failure becomes more apparent as more content is generated. This aligns with the idea that as more information is revealed, the model’s understanding of success or failure becomes clearer, similar to how humans typically gain better insight into a task as they progress further. We selected the layer that performed best on the given dataset in prior evaluations for the prediction-over-time experiments. Interestingly, in the \texttt{Llama-3.1-8B}-generated dataset, BERT’s performance tends to decline as more context is revealed, whereas for \texttt{Mistral-7B}, it remains relatively stable. While BERT can effectively process surface-level cues such as question structure, it appears less capable of tracking the evolving reasoning embedded in the chain-of-thought (CoT) generation. As the CoT unfolds and more tokens are added, the increasing complexity may exceed BERT’s shallow interpretive capacity, limiting its ability to follow deeper logical developments. In contrast, using internal LLM representations as input significantly improves performance—e.g., from 60\% to 69.4\% on AQuA in the \texttt{Llama-3.1-8B}-generated dataset—and yields consistent gains across all \texttt{Mistral-7B} datasets. For Olympiad and cn-k12 with \texttt{Llama-3.1-8B}, results remain comparable. This demonstrates the representations’ ability to capture more nuanced, cumulative reasoning and better reflect the underlying complexity of the CoT process, ultimately enabling more accurate success prediction. Surprisingly, \texttt{Mistral-7B}-generated AQuA dataset performs better as compared to the \texttt{Llama-3.1-8B}-generated one. Further inspection of the generations revealed that for the negative examples, \texttt{Mistral-7B} generations often assume that the provided question might be incorrect. These linguistic queues as shown in \autoref{tab:ling-ques-mistral-aqua}, might be used by BERT baseline as a feature to classify negative examples.

\begin{table*}
\small
\centering
\begin{tabular}{p{15.4cm}l}
\textbf{Example 1}\\
\toprule 
1. First, we need to find the probability that Kyle and Catherine solve the problem but David does not. \\
2. The probability that Kyle solves the problem is 1/3, and the probability that Catherine solves the problem is 5/9. \\
3. The probability that David does not solve the problem is 1 - 2/7 = 5/7. \\
4. To find the probability that all three events happen, we multiply their individual probabilities: (1/3) * (5/9) * (5/7). \\
5. Simplify the expression: (5/27) * (5/7) = 25/189. \\
6. \textbf{The answer is E) 20 / 189. However, there seems to be a mistake in the options provided. The correct answer should be 25/189, not 20/189. So, the correct answer is E) 25 / 189.} \\
\toprule

\textbf{Example 2}\\
\toprule 

1. The total investment is Rs. 70000 (A) + Rs. 120000 (B) = Rs. 190000. \\
2. The total profit is Rs. 26000. \\
3. To find the share of B, we need to divide the total profit by the total investment and then multiply it by B's investment. \\
Share of B = (Total profit / Total investment) * B's investment \\
= (26000 / 190000) * 120000 \\
= (26/19) * 12000 \\
= 24000 * (26/19) \\
= 24000 * (1.3158) \\
= 24009 \\

So, the share of B is approximately Rs. 24009. \\

\textbf{However, since the options provided are not exact, we can round the answer to the nearest option. The closest option is D) 24009. But since the question asks for one of the options, we cannot be certain that D) 24009 is the correct answer. It's a tricky question due to the rounding of the options.} \\

\textbf{Final answer: The share of B is approximately 24009, but the question asks for one of the options, so the exact answer cannot be determined from the given options.}\\

\toprule
\end{tabular}
\caption{Linguistic Patterns in the negative examples of \texttt{Mistral-7B}-generated AQuA dataset, which might be used by BERT classifier as a feature in our Prediction over Time experiments.}
\label{tab:ling-ques-mistral-aqua}
\end{table*}

\paragraph{Complimentary Analysis for Probing}
Our probing results suggest that the initial CoT steps in the Olympiad and Cn-K12 \texttt{Llama-3.1-8B} generated datasets contain similar knowledge about CoT success as the later stages, which may indicate that they share an understanding of the calculations performed up to that point. However, prior research has highlighted several limitations of probing. For instance, when trained on a large number of examples, models using randomly initialized embeddings as input representations can outperform those using pretrained intermediate representations \cite{zhang2018language, wieting2019no}.
To complement probing-based analyses, several alternative methods assess knowledge in model representations by measuring their correlations without relying on a classifier. These include SVCCA (Singular Vector Canonical Correlation Analysis) \cite{raghu2017svccasingularvectorcanonical}, PWCCA (Projection Weighted Canonical Correlation Analysis) \cite{morcos2018insights}, and CKA (Centered Kernel Alignment) \cite{kornblith2019similarity}, among others. Such methods are frequently used as alternatives for probing in NLP research \cite{saphra2018understanding,zhao-etal-2023-joint,zhao-etal-2024-layer}.
In our study, we employ SVCCA following \citet{zhao-etal-2022-understanding}, which investigated how domain-specific information is encoded in LLMs—a research focus related to ours.

We compute SVCCA scores between representations at each reasoning step\footnote{See \autoref{app:additional-results} for scores between all possible combinations of representations reasoning step} (10\%, 20\%, ..., 90\%) with 100\% (CoT completion), where higher scores indicate greater similarity in encoded information (\autoref{fig:pot_layer14_graph}). Since the AQuA dataset consists of shorter questions and CoT sequences (see \autoref{tab:ds_dist}), one might expect its intermediate representations to be more similar across steps, as the reasoning process is more compact. However, we observe that AQuA exhibits lower SVCCA scores between earlier and later steps compared to Olympiad and Cn-K12. This strengthens our probing findings: when earlier steps contain information predictive of CoT success, representations remain more stable throughout the reasoning process. This suggests that intermediate representations may not only encode predictive information about the final answer correctness but also the final answer itself. If the model is implicitly performing parts of the final computation at earlier steps, we may be able to leverage this by directly prompting it to provide an answer before completing the full chain of thought. 

\begin{table*}
    \centering
    \begin{tabular}{lcccccc}
        \toprule
        
        \textbf{Dataset} &  \textbf{Gen \%} & \textbf{Consistent} & \textbf{Inconsistent} & \textbf{Corrected} & \textbf{Correct} & \textbf{W\_CoT\_Correct}\\
        \midrule
        
        \multirow{3}{*}{AQuA}    & 50  & 40  & 45 & 15  & 37 & \multirow{3}{*}{38}\\
                                & 70 & 48  & 41 & 11  & 37&\\
                                & 99 & 81  & 14 & 5  & {59} & \\
        \midrule        
        \multirow{3}{*}{Olympiad}  & 30 & 21  & 76 & 3  & 18  & \multirow{3}{*}{19}\\
                                & 50 & 28  & 70 & 2  & 22 &\\
                                & 99 & 57  & 41 & 2  & 33 &\\
        \midrule        
        \multirow{3}{*}{Cn-k12}    & 30 & {37}  & {55} & {7}   & 32&\multirow{3}{*}{24}  \\
                                & 50 & {42}  & {49} & {7}  & 33 &\\
                                & 99 & {88}  & {9} & {3}  & 47 &\\

        \bottomrule
    \end{tabular}

\caption{LLM's generation was artificially interrupted to halt computation and were asked to just provide their best guess on 100 samples at 3 different time steps defined Gen. Annotators marked an answer Consistent if it is same the answer it provides when it is allowed to continue generation and Inconsistent if the provided answer differ. There were cases where full generation led to incorrect solutions by LLM and these interruptions allowed LLM to generate the correct answer, which was given the label Correct by annotators. }
\label{tab:human-annotation-interrupted-gen}
\end{table*}

\paragraph{Early Stopping in CoT Reasoning}
To examine whether the model's intermediate representations encode sufficient information for correct answers, we prompt it to generate an answer at various reasoning steps and evaluate its accuracy. As in our previous experiments, this intervention is performed in a zero-shot manner without explicitly training the model to follow such instructions. Specifically, we halt the \texttt{Llama-3.1-8B} reasoning process by providing the generated sequence up to a certain point, followed by the instruction: \emph{"Stop all computation and give me the correct answer in 2–3 words, if you already know it"}. This allows us to assess whether the model can extract a final answer without completing the full chain of thought (See Appendix \ref{app:early-stopping} for an illustration).
We conduct human annotation on 100 samples per dataset at three timesteps\footnote{We select two intervals where the classification model achieves its highest accuracy and one at the final step.}, summarizing the results in \autoref{tab:human-annotation-interrupted-gen}. Our analysis examines how often the halted response remains consistent with the final, uninterrupted answer, how often it changes (inconsistent), and in what fraction of those cases the change leads to a corrected answer—where stopping CoT reasoning actually improves performance. Finally, we compare overall correctness rates to both the full CoT process and a setting without CoT.
The relatively low consistency rate, even at 99\% completion—particularly in Olympiad (57\%)—suggests that zero-shot early stopping is a suboptimal intervention. The model does not always converge to a stable answer, even when nearly the entire reasoning sequence is generated, highlighting the limitations of simply prompting it to stop early. However, despite this brittleness, halting CoT midway in AQuA and Cn-K12—the two datasets where later reasoning steps did not enhance CoT success predictability—still slightly outperforms the setting without CoT. This indicates that even incomplete CoT sequences can carry enough information to improve accuracy, revealing untapped potential in intermediate reasoning states.
These findings suggest that while zero-shot interventions have limitations, more sophisticated approaches—such as training the model to generate concise reasoning chains through supervised learning or reinforcing brevity via RL-based rewards—could more effectively unlock this potential. By optimizing the model's ability to extract key reasoning steps without unnecessary verbosity, future methods could further bridge the gap between full CoT and early stopping while maintaining or even improving accuracy.\footnote{In some cases, halting CoT before completion improved answer correctness, as reflected in the \emph{corrected} column in the table.}



\section{Conclusion}
We demonstrate that the success of the CoT reasoning process can be predicted from the internal representations of the LLM even before the generation of a single token. However, we also observe that, in some cases, the accuracy of this prediction does not improve when the classifier is exposed to intermediate reasoning steps. Using SVCCA, we show that early steps encode information that is more similar to the final steps in these cases. This raises the question of whether these early representations also contain valuable information about the final answer itself. Our initial experiments suggest that while this potential exists, zero-shot prompting may not fully unlock it. We hope our findings will inform future research aimed at making CoT more efficient without sacrificing accuracy, as the computational cost of CoT is significant.

\section*{Limitation}
Manually annotating test examples for each dataset limits the generalizability of our study, as we use human evaluation for only a single LLM and focus on three math datasets in a zero-shot setting. Additionally, we use a temperature of zero to minimize stochastic noise in the generation process, which could accumulate over multiple reasoning steps. However, this setting may not fully capture the variability present in real-world, stochastic LLM usage. Furthermore, our method assumes white-box access to the model, which is not typically available for proprietary models.

\section*{Ethics Statement} Our research focused on evaluating the internal representation of a Large Language Model to better the notion of prediction for CoT Reasoning. During out research, we didn't perform any fine-tuning that could introduce any bias in the LLM. We work with open-source datasets and hence produce no additional bias than what might already be part of it. During our manual evaluations, annotators were asked to correct math problems, where there is only one logically correct answer, reducing the suspect of biased annotations. We mostly used existing algorithms for supporting our analysis, hence making our findings more reliable. For reproducibility, we release the hyper-parameters used in our experiments.
\bibliography{custom}

\begin{thebibliography}{40}
\providecommand{\natexlab}[1]{#1}

\bibitem[{Ahn et~al.(2024)Ahn, Verma, Lou, Liu, Zhang, and Yin}]{ahn-etal-2024-large}
Janice Ahn, Rishu Verma, Renze Lou, Di~Liu, Rui Zhang, and Wenpeng Yin. 2024.
\newblock \href {https://aclanthology.org/2024.eacl-srw.17/} {Large language models for mathematical reasoning: Progresses and challenges}.
\newblock In \emph{Proceedings of the 18th Conference of the European Chapter of the Association for Computational Linguistics: Student Research Workshop}, pages 225--237, St. Julian{'}s, Malta. Association for Computational Linguistics.

\bibitem[{Azaria and Mitchell(2023)}]{azaria-mitchell-2023-internal}
Amos Azaria and Tom Mitchell. 2023.
\newblock \href {https://doi.org/10.18653/v1/2023.findings-emnlp.68} {The internal state of an {LLM} knows when it`s lying}.
\newblock In \emph{Findings of the Association for Computational Linguistics: EMNLP 2023}, pages 967--976, Singapore. Association for Computational Linguistics.

\bibitem[{Bao et~al.(2025)Bao, Zhang, Wang, Yang, and Zhang}]{bao-etal-2025-likely}
Guangsheng Bao, Hongbo Zhang, Cunxiang Wang, Linyi Yang, and Yue Zhang. 2025.
\newblock \href {https://aclanthology.org/2025.coling-main.524/} {How likely do {LLM}s with {C}o{T} mimic human reasoning?}
\newblock In \emph{Proceedings of the 31st International Conference on Computational Linguistics}, pages 7831--7850, Abu Dhabi, UAE. Association for Computational Linguistics.

\bibitem[{Belinkov(2022)}]{belinkov2022probing}
Yonatan Belinkov. 2022.
\newblock Probing classifiers: Promises, shortcomings, and advances.
\newblock \emph{Computational Linguistics}, 48(1):207--219.

\bibitem[{Byun et~al.(2024)Byun, Chun, Kil, and Perrault}]{byun-etal-2024-ares}
Ju-Seung Byun, Jiyun Chun, Jihyung Kil, and Andrew Perrault. 2024.
\newblock \href {https://doi.org/10.18653/v1/2024.emnlp-main.252} {{ARES}: Alternating reinforcement learning and supervised fine-tuning for enhanced multi-modal chain-of-thought reasoning through diverse {AI} feedback}.
\newblock In \emph{Proceedings of the 2024 Conference on Empirical Methods in Natural Language Processing}, pages 4410--4430, Miami, Florida, USA. Association for Computational Linguistics.

\bibitem[{Cui et~al.(2024)Cui, He, Tang, He, Luo, Tang, and Xing}]{cui2024theoreticalunderstandingchainofthoughtcoherent}
Yingqian Cui, Pengfei He, Xianfeng Tang, Qi~He, Chen Luo, Jiliang Tang, and Yue Xing. 2024.
\newblock \href {https://arxiv.org/abs/2410.16540} {A theoretical understanding of chain-of-thought: Coherent reasoning and error-aware demonstration}.
\newblock \emph{Preprint}, arXiv:2410.16540.

\bibitem[{Devlin et~al.(2019)Devlin, Chang, Lee, and Toutanova}]{devlin-etal-2019-bert}
Jacob Devlin, Ming-Wei Chang, Kenton Lee, and Kristina Toutanova. 2019.
\newblock \href {https://doi.org/10.18653/v1/N19-1423} {{BERT}: Pre-training of deep bidirectional transformers for language understanding}.
\newblock In \emph{Proceedings of the 2019 Conference of the North {A}merican Chapter of the Association for Computational Linguistics: Human Language Technologies, Volume 1 (Long and Short Papers)}, pages 4171--4186, Minneapolis, Minnesota. Association for Computational Linguistics.

\bibitem[{Gottesman and Geva(2024)}]{gottesman-geva-2024-estimating}
Daniela Gottesman and Mor Geva. 2024.
\newblock \href {https://doi.org/10.18653/v1/2024.emnlp-main.232} {Estimating knowledge in large language models without generating a single token}.
\newblock In \emph{Proceedings of the 2024 Conference on Empirical Methods in Natural Language Processing}, pages 3994--4019, Miami, Florida, USA. Association for Computational Linguistics.

\bibitem[{Goyal et~al.(2024)Goyal, Ji, Rawat, Menon, Kumar, and Nagarajan}]{goyal2024thinkspeaktraininglanguage}
Sachin Goyal, Ziwei Ji, Ankit~Singh Rawat, Aditya~Krishna Menon, Sanjiv Kumar, and Vaishnavh Nagarajan. 2024.
\newblock \href {https://arxiv.org/abs/2310.02226} {Think before you speak: Training language models with pause tokens}.
\newblock \emph{Preprint}, arXiv:2310.02226.

\bibitem[{Grattafiori et~al.(2024)Grattafiori, Dubey, Jauhri, Pandey, Kadian, Al-Dahle, Letman, Mathur, Schelten, Vaughan, Yang, Fan, Goyal, Hartshorn, Yang, Mitra, Sravankumar, Korenev, Hinsvark, Rao, Zhang, Rodriguez, Gregerson, Spataru, Roziere, Biron, Tang, Chern, Caucheteux, Nayak, Bi, Marra, McConnell, Keller, Touret, Wu, Wong, Ferrer, Nikolaidis, Allonsius, Song, Pintz, Livshits, Wyatt, Esiobu, Choudhary, Mahajan, Garcia-Olano, Perino, Hupkes, Lakomkin, AlBadawy, Lobanova, Dinan, Smith, Radenovic, Guzmán, Zhang, Synnaeve, Lee, Anderson, Thattai, Nail, Mialon, Pang, Cucurell, Nguyen, Korevaar, Xu, Touvron, Zarov, Ibarra, Kloumann, Misra, Evtimov, Zhang, Copet, Lee, Geffert, Vranes, Park, Mahadeokar, Shah, van~der Linde, Billock, Hong, Lee, Fu, Chi, Huang, Liu, Wang, Yu, Bitton, Spisak, Park, Rocca, Johnstun, Saxe, Jia, Alwala, Prasad, Upasani, Plawiak, Li, Heafield, Stone, El-Arini, Iyer, Malik, Chiu, Bhalla, Lakhotia, Rantala-Yeary, van~der Maaten, Chen, Tan, Jenkins, Martin, Madaan, Malo, Blecher,
  Landzaat, de~Oliveira, Muzzi, Pasupuleti, Singh, Paluri, Kardas, Tsimpoukelli, Oldham, Rita, Pavlova, Kambadur, Lewis, Si, Singh, Hassan, Goyal, Torabi, Bashlykov, Bogoychev, Chatterji, Zhang, Duchenne, Çelebi, Alrassy, Zhang, Li, Vasic, Weng, Bhargava, Dubal, Krishnan, Koura, Xu, He, Dong, Srinivasan, Ganapathy, Calderer, Cabral, Stojnic, Raileanu, Maheswari, Girdhar, Patel, Sauvestre, Polidoro, Sumbaly, Taylor, Silva, Hou, Wang, Hosseini, Chennabasappa, Singh, Bell, Kim, Edunov, Nie, Narang, Raparthy, Shen, Wan, Bhosale, Zhang, Vandenhende, Batra, Whitman, Sootla, Collot, Gururangan, Borodinsky, Herman, Fowler, Sheasha, Georgiou, Scialom, Speckbacher, Mihaylov, Xiao, Karn, Goswami, Gupta, Ramanathan, Kerkez, Gonguet, Do, Vogeti, Albiero, Petrovic, Chu, Xiong, Fu, Meers, Martinet, Wang, Wang, Tan, Xia, Xie, Jia, Wang, Goldschlag, Gaur, Babaei, Wen, Song, Zhang, Li, Mao, Coudert, Yan, Chen, Papakipos, Singh, Srivastava, Jain, Kelsey, Shajnfeld, Gangidi, Victoria, Goldstand, Menon, Sharma, Boesenberg,
  Baevski, Feinstein, Kallet, Sangani, Teo, Yunus, Lupu, Alvarado, Caples, Gu, Ho, Poulton, Ryan, Ramchandani, Dong, Franco, Goyal, Saraf, Chowdhury, Gabriel, Bharambe, Eisenman, Yazdan, James, Maurer, Leonhardi, Huang, Loyd, Paola, Paranjape, Liu, Wu, Ni, Hancock, Wasti, Spence, Stojkovic, Gamido, Montalvo, Parker, Burton, Mejia, Liu, Wang, Kim, Zhou, Hu, Chu, Cai, Tindal, Feichtenhofer, Gao, Civin, Beaty, Kreymer, Li, Adkins, Xu, Testuggine, David, Parikh, Liskovich, Foss, Wang, Le, Holland, Dowling, Jamil, Montgomery, Presani, Hahn, Wood, Le, Brinkman, Arcaute, Dunbar, Smothers, Sun, Kreuk, Tian, Kokkinos, Ozgenel, Caggioni, Kanayet, Seide, Florez, Schwarz, Badeer, Swee, Halpern, Herman, Sizov, Guangyi, Zhang, Lakshminarayanan, Inan, Shojanazeri, Zou, Wang, Zha, Habeeb, Rudolph, Suk, Aspegren, Goldman, Zhan, Damlaj, Molybog, Tufanov, Leontiadis, Veliche, Gat, Weissman, Geboski, Kohli, Lam, Asher, Gaya, Marcus, Tang, Chan, Zhen, Reizenstein, Teboul, Zhong, Jin, Yang, Cummings, Carvill, Shepard, McPhie,
  Torres, Ginsburg, Wang, Wu, U, Saxena, Khandelwal, Zand, Matosich, Veeraraghavan, Michelena, Li, Jagadeesh, Huang, Chawla, Huang, Chen, Garg, A, Silva, Bell, Zhang, Guo, Yu, Moshkovich, Wehrstedt, Khabsa, Avalani, Bhatt, Mankus, Hasson, Lennie, Reso, Groshev, Naumov, Lathi, Keneally, Liu, Seltzer, Valko, Restrepo, Patel, Vyatskov, Samvelyan, Clark, Macey, Wang, Hermoso, Metanat, Rastegari, Bansal, Santhanam, Parks, White, Bawa, Singhal, Egebo, Usunier, Mehta, Laptev, Dong, Cheng, Chernoguz, Hart, Salpekar, Kalinli, Kent, Parekh, Saab, Balaji, Rittner, Bontrager, Roux, Dollar, Zvyagina, Ratanchandani, Yuvraj, Liang, Alao, Rodriguez, Ayub, Murthy, Nayani, Mitra, Parthasarathy, Li, Hogan, Battey, Wang, Howes, Rinott, Mehta, Siby, Bondu, Datta, Chugh, Hunt, Dhillon, Sidorov, Pan, Mahajan, Verma, Yamamoto, Ramaswamy, Lindsay, Lindsay, Feng, Lin, Zha, Patil, Shankar, Zhang, Zhang, Wang, Agarwal, Sajuyigbe, Chintala, Max, Chen, Kehoe, Satterfield, Govindaprasad, Gupta, Deng, Cho, Virk, Subramanian, Choudhury,
  Goldman, Remez, Glaser, Best, Koehler, Robinson, Li, Zhang, Matthews, Chou, Shaked, Vontimitta, Ajayi, Montanez, Mohan, Kumar, Mangla, Ionescu, Poenaru, Mihailescu, Ivanov, Li, Wang, Jiang, Bouaziz, Constable, Tang, Wu, Wang, Wu, Gao, Kleinman, Chen, Hu, Jia, Qi, Li, Zhang, Zhang, Adi, Nam, Yu, Wang, Zhao, Hao, Qian, Li, He, Rait, DeVito, Rosnbrick, Wen, Yang, Zhao, and Ma}]{grattafiori2024llama3herdmodels}
Aaron Grattafiori, Abhimanyu Dubey, Abhinav Jauhri, Abhinav Pandey, Abhishek Kadian, Ahmad Al-Dahle, Aiesha Letman, Akhil Mathur, Alan Schelten, Alex Vaughan, Amy Yang, Angela Fan, Anirudh Goyal, Anthony Hartshorn, Aobo Yang, Archi Mitra, Archie Sravankumar, Artem Korenev, Arthur Hinsvark, Arun Rao, Aston Zhang, Aurelien Rodriguez, Austen Gregerson, Ava Spataru, Baptiste Roziere, Bethany Biron, Binh Tang, Bobbie Chern, Charlotte Caucheteux, Chaya Nayak, Chloe Bi, Chris Marra, Chris McConnell, Christian Keller, Christophe Touret, Chunyang Wu, Corinne Wong, Cristian~Canton Ferrer, Cyrus Nikolaidis, Damien Allonsius, Daniel Song, Danielle Pintz, Danny Livshits, Danny Wyatt, David Esiobu, Dhruv Choudhary, Dhruv Mahajan, Diego Garcia-Olano, Diego Perino, Dieuwke Hupkes, Egor Lakomkin, Ehab AlBadawy, Elina Lobanova, Emily Dinan, Eric~Michael Smith, Filip Radenovic, Francisco Guzmán, Frank Zhang, Gabriel Synnaeve, Gabrielle Lee, Georgia~Lewis Anderson, Govind Thattai, Graeme Nail, Gregoire Mialon, Guan Pang,
  Guillem Cucurell, Hailey Nguyen, Hannah Korevaar, Hu~Xu, Hugo Touvron, Iliyan Zarov, Imanol~Arrieta Ibarra, Isabel Kloumann, Ishan Misra, Ivan Evtimov, Jack Zhang, Jade Copet, Jaewon Lee, Jan Geffert, Jana Vranes, Jason Park, Jay Mahadeokar, Jeet Shah, Jelmer van~der Linde, Jennifer Billock, Jenny Hong, Jenya Lee, Jeremy Fu, Jianfeng Chi, Jianyu Huang, Jiawen Liu, Jie Wang, Jiecao Yu, Joanna Bitton, Joe Spisak, Jongsoo Park, Joseph Rocca, Joshua Johnstun, Joshua Saxe, Junteng Jia, Kalyan~Vasuden Alwala, Karthik Prasad, Kartikeya Upasani, Kate Plawiak, Ke~Li, Kenneth Heafield, Kevin Stone, Khalid El-Arini, Krithika Iyer, Kshitiz Malik, Kuenley Chiu, Kunal Bhalla, Kushal Lakhotia, Lauren Rantala-Yeary, Laurens van~der Maaten, Lawrence Chen, Liang Tan, Liz Jenkins, Louis Martin, Lovish Madaan, Lubo Malo, Lukas Blecher, Lukas Landzaat, Luke de~Oliveira, Madeline Muzzi, Mahesh Pasupuleti, Mannat Singh, Manohar Paluri, Marcin Kardas, Maria Tsimpoukelli, Mathew Oldham, Mathieu Rita, Maya Pavlova, Melanie Kambadur,
  Mike Lewis, Min Si, Mitesh~Kumar Singh, Mona Hassan, Naman Goyal, Narjes Torabi, Nikolay Bashlykov, Nikolay Bogoychev, Niladri Chatterji, Ning Zhang, Olivier Duchenne, Onur Çelebi, Patrick Alrassy, Pengchuan Zhang, Pengwei Li, Petar Vasic, Peter Weng, Prajjwal Bhargava, Pratik Dubal, Praveen Krishnan, Punit~Singh Koura, Puxin Xu, Qing He, Qingxiao Dong, Ragavan Srinivasan, Raj Ganapathy, Ramon Calderer, Ricardo~Silveira Cabral, Robert Stojnic, Roberta Raileanu, Rohan Maheswari, Rohit Girdhar, Rohit Patel, Romain Sauvestre, Ronnie Polidoro, Roshan Sumbaly, Ross Taylor, Ruan Silva, Rui Hou, Rui Wang, Saghar Hosseini, Sahana Chennabasappa, Sanjay Singh, Sean Bell, Seohyun~Sonia Kim, Sergey Edunov, Shaoliang Nie, Sharan Narang, Sharath Raparthy, Sheng Shen, Shengye Wan, Shruti Bhosale, Shun Zhang, Simon Vandenhende, Soumya Batra, Spencer Whitman, Sten Sootla, Stephane Collot, Suchin Gururangan, Sydney Borodinsky, Tamar Herman, Tara Fowler, Tarek Sheasha, Thomas Georgiou, Thomas Scialom, Tobias Speckbacher,
  Todor Mihaylov, Tong Xiao, Ujjwal Karn, Vedanuj Goswami, Vibhor Gupta, Vignesh Ramanathan, Viktor Kerkez, Vincent Gonguet, Virginie Do, Vish Vogeti, Vítor Albiero, Vladan Petrovic, Weiwei Chu, Wenhan Xiong, Wenyin Fu, Whitney Meers, Xavier Martinet, Xiaodong Wang, Xiaofang Wang, Xiaoqing~Ellen Tan, Xide Xia, Xinfeng Xie, Xuchao Jia, Xuewei Wang, Yaelle Goldschlag, Yashesh Gaur, Yasmine Babaei, Yi~Wen, Yiwen Song, Yuchen Zhang, Yue Li, Yuning Mao, Zacharie~Delpierre Coudert, Zheng Yan, Zhengxing Chen, Zoe Papakipos, Aaditya Singh, Aayushi Srivastava, Abha Jain, Adam Kelsey, Adam Shajnfeld, Adithya Gangidi, Adolfo Victoria, Ahuva Goldstand, Ajay Menon, Ajay Sharma, Alex Boesenberg, Alexei Baevski, Allie Feinstein, Amanda Kallet, Amit Sangani, Amos Teo, Anam Yunus, Andrei Lupu, Andres Alvarado, Andrew Caples, Andrew Gu, Andrew Ho, Andrew Poulton, Andrew Ryan, Ankit Ramchandani, Annie Dong, Annie Franco, Anuj Goyal, Aparajita Saraf, Arkabandhu Chowdhury, Ashley Gabriel, Ashwin Bharambe, Assaf Eisenman, Azadeh
  Yazdan, Beau James, Ben Maurer, Benjamin Leonhardi, Bernie Huang, Beth Loyd, Beto~De Paola, Bhargavi Paranjape, Bing Liu, Bo~Wu, Boyu Ni, Braden Hancock, Bram Wasti, Brandon Spence, Brani Stojkovic, Brian Gamido, Britt Montalvo, Carl Parker, Carly Burton, Catalina Mejia, Ce~Liu, Changhan Wang, Changkyu Kim, Chao Zhou, Chester Hu, Ching-Hsiang Chu, Chris Cai, Chris Tindal, Christoph Feichtenhofer, Cynthia Gao, Damon Civin, Dana Beaty, Daniel Kreymer, Daniel Li, David Adkins, David Xu, Davide Testuggine, Delia David, Devi Parikh, Diana Liskovich, Didem Foss, Dingkang Wang, Duc Le, Dustin Holland, Edward Dowling, Eissa Jamil, Elaine Montgomery, Eleonora Presani, Emily Hahn, Emily Wood, Eric-Tuan Le, Erik Brinkman, Esteban Arcaute, Evan Dunbar, Evan Smothers, Fei Sun, Felix Kreuk, Feng Tian, Filippos Kokkinos, Firat Ozgenel, Francesco Caggioni, Frank Kanayet, Frank Seide, Gabriela~Medina Florez, Gabriella Schwarz, Gada Badeer, Georgia Swee, Gil Halpern, Grant Herman, Grigory Sizov, Guangyi, Zhang, Guna
  Lakshminarayanan, Hakan Inan, Hamid Shojanazeri, Han Zou, Hannah Wang, Hanwen Zha, Haroun Habeeb, Harrison Rudolph, Helen Suk, Henry Aspegren, Hunter Goldman, Hongyuan Zhan, Ibrahim Damlaj, Igor Molybog, Igor Tufanov, Ilias Leontiadis, Irina-Elena Veliche, Itai Gat, Jake Weissman, James Geboski, James Kohli, Janice Lam, Japhet Asher, Jean-Baptiste Gaya, Jeff Marcus, Jeff Tang, Jennifer Chan, Jenny Zhen, Jeremy Reizenstein, Jeremy Teboul, Jessica Zhong, Jian Jin, Jingyi Yang, Joe Cummings, Jon Carvill, Jon Shepard, Jonathan McPhie, Jonathan Torres, Josh Ginsburg, Junjie Wang, Kai Wu, Kam~Hou U, Karan Saxena, Kartikay Khandelwal, Katayoun Zand, Kathy Matosich, Kaushik Veeraraghavan, Kelly Michelena, Keqian Li, Kiran Jagadeesh, Kun Huang, Kunal Chawla, Kyle Huang, Lailin Chen, Lakshya Garg, Lavender A, Leandro Silva, Lee Bell, Lei Zhang, Liangpeng Guo, Licheng Yu, Liron Moshkovich, Luca Wehrstedt, Madian Khabsa, Manav Avalani, Manish Bhatt, Martynas Mankus, Matan Hasson, Matthew Lennie, Matthias Reso, Maxim
  Groshev, Maxim Naumov, Maya Lathi, Meghan Keneally, Miao Liu, Michael~L. Seltzer, Michal Valko, Michelle Restrepo, Mihir Patel, Mik Vyatskov, Mikayel Samvelyan, Mike Clark, Mike Macey, Mike Wang, Miquel~Jubert Hermoso, Mo~Metanat, Mohammad Rastegari, Munish Bansal, Nandhini Santhanam, Natascha Parks, Natasha White, Navyata Bawa, Nayan Singhal, Nick Egebo, Nicolas Usunier, Nikhil Mehta, Nikolay~Pavlovich Laptev, Ning Dong, Norman Cheng, Oleg Chernoguz, Olivia Hart, Omkar Salpekar, Ozlem Kalinli, Parkin Kent, Parth Parekh, Paul Saab, Pavan Balaji, Pedro Rittner, Philip Bontrager, Pierre Roux, Piotr Dollar, Polina Zvyagina, Prashant Ratanchandani, Pritish Yuvraj, Qian Liang, Rachad Alao, Rachel Rodriguez, Rafi Ayub, Raghotham Murthy, Raghu Nayani, Rahul Mitra, Rangaprabhu Parthasarathy, Raymond Li, Rebekkah Hogan, Robin Battey, Rocky Wang, Russ Howes, Ruty Rinott, Sachin Mehta, Sachin Siby, Sai~Jayesh Bondu, Samyak Datta, Sara Chugh, Sara Hunt, Sargun Dhillon, Sasha Sidorov, Satadru Pan, Saurabh Mahajan,
  Saurabh Verma, Seiji Yamamoto, Sharadh Ramaswamy, Shaun Lindsay, Shaun Lindsay, Sheng Feng, Shenghao Lin, Shengxin~Cindy Zha, Shishir Patil, Shiva Shankar, Shuqiang Zhang, Shuqiang Zhang, Sinong Wang, Sneha Agarwal, Soji Sajuyigbe, Soumith Chintala, Stephanie Max, Stephen Chen, Steve Kehoe, Steve Satterfield, Sudarshan Govindaprasad, Sumit Gupta, Summer Deng, Sungmin Cho, Sunny Virk, Suraj Subramanian, Sy~Choudhury, Sydney Goldman, Tal Remez, Tamar Glaser, Tamara Best, Thilo Koehler, Thomas Robinson, Tianhe Li, Tianjun Zhang, Tim Matthews, Timothy Chou, Tzook Shaked, Varun Vontimitta, Victoria Ajayi, Victoria Montanez, Vijai Mohan, Vinay~Satish Kumar, Vishal Mangla, Vlad Ionescu, Vlad Poenaru, Vlad~Tiberiu Mihailescu, Vladimir Ivanov, Wei Li, Wenchen Wang, Wenwen Jiang, Wes Bouaziz, Will Constable, Xiaocheng Tang, Xiaojian Wu, Xiaolan Wang, Xilun Wu, Xinbo Gao, Yaniv Kleinman, Yanjun Chen, Ye~Hu, Ye~Jia, Ye~Qi, Yenda Li, Yilin Zhang, Ying Zhang, Yossi Adi, Youngjin Nam, Yu, Wang, Yu~Zhao, Yuchen Hao, Yundi
  Qian, Yunlu Li, Yuzi He, Zach Rait, Zachary DeVito, Zef Rosnbrick, Zhaoduo Wen, Zhenyu Yang, Zhiwei Zhao, and Zhiyu Ma. 2024.
\newblock \href {https://arxiv.org/abs/2407.21783} {The llama 3 herd of models}.
\newblock \emph{Preprint}, arXiv:2407.21783.

\bibitem[{Ji et~al.(2025)Ji, Song, Zhong, Jia, Wu, Cao, and Xu}]{ji2025mygomultiplexcotmethod}
Shihao Ji, Zihui Song, Fucheng Zhong, Jisen Jia, Zhaobo Wu, Zheyi Cao, and Tianhao Xu. 2025.
\newblock \href {https://arxiv.org/abs/2501.13117} {Mygo multiplex cot: A method for self-reflection in large language models via double chain of thought thinking}.
\newblock \emph{Preprint}, arXiv:2501.13117.

\bibitem[{Kornblith et~al.(2019)Kornblith, Norouzi, Lee, and Hinton}]{kornblith2019similarity}
Simon Kornblith, Mohammad Norouzi, Honglak Lee, and Geoffrey Hinton. 2019.
\newblock Similarity of neural network representations revisited.
\newblock In \emph{International conference on machine learning}, pages 3519--3529. PMLR.

\bibitem[{Kumari et~al.(2024)Kumari, Jain, and Ekbal}]{kumari-etal-2024-m3hop}
Gitanjali Kumari, Kirtan Jain, and Asif Ekbal. 2024.
\newblock \href {https://doi.org/10.18653/v1/2024.emnlp-main.1234} {{M}3{H}op-{C}o{T}: Misogynous meme identification with multimodal multi-hop chain-of-thought}.
\newblock In \emph{Proceedings of the 2024 Conference on Empirical Methods in Natural Language Processing}, pages 22105--22138, Miami, Florida, USA. Association for Computational Linguistics.

\bibitem[{Li et~al.(2023)Li, Hammoud, Itani, Khizbullin, and Ghanem}]{li2023camel}
Guohao Li, Hasan Abed Al~Kader Hammoud, Hani Itani, Dmitrii Khizbullin, and Bernard Ghanem. 2023.
\newblock \href {https://arxiv.org/abs/2303.17760} {Camel: Communicative agents for "mind" exploration of large scale language model society}.
\newblock \emph{Preprint}, arXiv:2303.17760.

\bibitem[{LI et~al.(2024)LI, Beeching, Tunstall, Lipkin, Soletskyi, Huang, Rasul, Yu, Jiang, Shen, Qin, Dong, Zhou, Fleureau, Lample, and Polu}]{numina_math_datasets}
Jia LI, Edward Beeching, Lewis Tunstall, Ben Lipkin, Roman Soletskyi, Shengyi~Costa Huang, Kashif Rasul, Longhui Yu, Albert Jiang, Ziju Shen, Zihan Qin, Bin Dong, Li~Zhou, Yann Fleureau, Guillaume Lample, and Stanislas Polu. 2024.
\newblock Numinamath.
\newblock \url{[https://huggingface.co/AI-MO/NuminaMath-CoT](https://github.com/project-numina/aimo-progress-prize/blob/main/report/numina_dataset.pdf)}.

\bibitem[{Ling et~al.(2017)Ling, Yogatama, Dyer, and Blunsom}]{ling2017program}
Wang Ling, Dani Yogatama, Chris Dyer, and Phil Blunsom. 2017.
\newblock Program induction by rationale generation: Learning to solve and explain algebraic word problems.
\newblock \emph{ACL}.

\bibitem[{Liu et~al.(2024{\natexlab{a}})Liu, Wang, Du, Zhou, and Liu}]{liu-etal-2024-medcot}
Jiaxiang Liu, Yuan Wang, Jiawei Du, Joey~Tianyi Zhou, and Zuozhu Liu. 2024{\natexlab{a}}.
\newblock \href {https://doi.org/10.18653/v1/2024.emnlp-main.962} {{M}ed{C}o{T}: Medical chain of thought via hierarchical expert}.
\newblock In \emph{Proceedings of the 2024 Conference on Empirical Methods in Natural Language Processing}, pages 17371--17389, Miami, Florida, USA. Association for Computational Linguistics.

\bibitem[{Liu et~al.(2024{\natexlab{b}})Liu, Geng, Wu, Sucholutsky, Lombrozo, and Griffiths}]{liu2024mindstepbystep}
Ryan Liu, Jiayi Geng, Addison~J. Wu, Ilia Sucholutsky, Tania Lombrozo, and Thomas~L. Griffiths. 2024{\natexlab{b}}.
\newblock \href {https://arxiv.org/abs/2410.21333} {Mind your step (by step): Chain-of-thought can reduce performance on tasks where thinking makes humans worse}.
\newblock \emph{Preprint}, arXiv:2410.21333.

\bibitem[{Madaan and Yazdanbakhsh(2022)}]{madaan2022text}
Aman Madaan and Amir Yazdanbakhsh. 2022.
\newblock Text and patterns: For effective chain of thought, it takes two to tango.
\newblock \emph{arXiv preprint arXiv:2209.07686}.

\bibitem[{Morcos et~al.(2018)Morcos, Raghu, and Bengio}]{morcos2018insights}
Ari Morcos, Maithra Raghu, and Samy Bengio. 2018.
\newblock Insights on representational similarity in neural networks with canonical correlation.
\newblock \emph{Advances in neural information processing systems}, 31.

\bibitem[{Pfau et~al.(2024)Pfau, Merrill, and Bowman}]{pfau2024let}
Jacob Pfau, William Merrill, and Samuel~R Bowman. 2024.
\newblock Let's think dot by dot: Hidden computation in transformer language models.
\newblock \emph{arXiv preprint arXiv:2404.15758}.

\bibitem[{Raghu et~al.(2017)Raghu, Gilmer, Yosinski, and Sohl-Dickstein}]{raghu2017svccasingularvectorcanonical}
Maithra Raghu, Justin Gilmer, Jason Yosinski, and Jascha Sohl-Dickstein. 2017.
\newblock \href {https://arxiv.org/abs/1706.05806} {Svcca: Singular vector canonical correlation analysis for deep learning dynamics and interpretability}.
\newblock \emph{Preprint}, arXiv:1706.05806.

\bibitem[{Rai and Yao(2024)}]{rai-yao-2024-investigation}
Daking Rai and Ziyu Yao. 2024.
\newblock \href {https://doi.org/10.18653/v1/2024.acl-long.387} {An investigation of neuron activation as a unified lens to explain chain-of-thought eliciting arithmetic reasoning of {LLM}s}.
\newblock In \emph{Proceedings of the 62nd Annual Meeting of the Association for Computational Linguistics (Volume 1: Long Papers)}, pages 7174--7193, Bangkok, Thailand. Association for Computational Linguistics.

\bibitem[{Renze(2024)}]{renze-2024-effect}
Matthew Renze. 2024.
\newblock \href {https://doi.org/10.18653/v1/2024.findings-emnlp.432} {The effect of sampling temperature on problem solving in large language models}.
\newblock In \emph{Findings of the Association for Computational Linguistics: EMNLP 2024}, pages 7346--7356, Miami, Florida, USA. Association for Computational Linguistics.

\bibitem[{Saphra and Lopez(2018)}]{saphra2018understanding}
Naomi Saphra and Adam Lopez. 2018.
\newblock Understanding learning dynamics of language models with svcca.
\newblock \emph{arXiv preprint arXiv:1811.00225}.

\bibitem[{Seo et~al.(2025)Seo, Lee, and Yeo}]{seodetecting}
Yeongbin Seo, Dongha Lee, and Jinyoung Yeo. 2025.
\newblock Detecting hallucination before answering: Semantic compression through instruction.

\bibitem[{Sprague et~al.(2024)Sprague, Yin, Rodriguez, Jiang, Wadhwa, Singhal, Zhao, Ye, Mahowald, and Durrett}]{sprague2024cotcotchainofthoughthelps}
Zayne Sprague, Fangcong Yin, Juan~Diego Rodriguez, Dongwei Jiang, Manya Wadhwa, Prasann Singhal, Xinyu Zhao, Xi~Ye, Kyle Mahowald, and Greg Durrett. 2024.
\newblock \href {https://arxiv.org/abs/2409.12183} {To cot or not to cot? chain-of-thought helps mainly on math and symbolic reasoning}.
\newblock \emph{Preprint}, arXiv:2409.12183.

\bibitem[{Turpin et~al.(2023)Turpin, Michael, Perez, and Bowman}]{NEURIPS2023_ed3fea90}
Miles Turpin, Julian Michael, Ethan Perez, and Samuel Bowman. 2023.
\newblock \href {https://proceedings.neurips.cc/paper_files/paper/2023/file/ed3fea9033a80fea1376299fa7863f4a-Paper-Conference.pdf} {Language models don\textquotesingle t always say what they think: Unfaithful explanations in chain-of-thought prompting}.
\newblock In \emph{Advances in Neural Information Processing Systems}, volume~36, pages 74952--74965. Curran Associates, Inc.

\bibitem[{Wang et~al.(2023{\natexlab{a}})Wang, Min, Deng, Shen, Wu, Zettlemoyer, and Sun}]{wang-etal-2023-towards}
Boshi Wang, Sewon Min, Xiang Deng, Jiaming Shen, You Wu, Luke Zettlemoyer, and Huan Sun. 2023{\natexlab{a}}.
\newblock \href {https://doi.org/10.18653/v1/2023.acl-long.153} {Towards understanding chain-of-thought prompting: An empirical study of what matters}.
\newblock In \emph{Proceedings of the 61st Annual Meeting of the Association for Computational Linguistics (Volume 1: Long Papers)}, pages 2717--2739, Toronto, Canada. Association for Computational Linguistics.

\bibitem[{Wang et~al.(2023{\natexlab{b}})Wang, Ren, Zhou, Lu, Luo, Shi, Zhang, Song, Zhan, and Li}]{wang2023mathcoderseamlesscodeintegration}
Ke~Wang, Houxing Ren, Aojun Zhou, Zimu Lu, Sichun Luo, Weikang Shi, Renrui Zhang, Linqi Song, Mingjie Zhan, and Hongsheng Li. 2023{\natexlab{b}}.
\newblock \href {https://arxiv.org/abs/2310.03731} {Mathcoder: Seamless code integration in llms for enhanced mathematical reasoning}.
\newblock \emph{Preprint}, arXiv:2310.03731.

\bibitem[{Wang et~al.(2024)Wang, Zeng, Zheng, Xing, Xu, and Xu}]{wang-etal-2024-videocot}
Yan Wang, Yawen Zeng, Jingsheng Zheng, Xiaofen Xing, Jin Xu, and Xiangmin Xu. 2024.
\newblock \href {https://doi.org/10.18653/v1/2024.alvr-1.8} {{V}ideo{C}o{T}: A video chain-of-thought dataset with active annotation tool}.
\newblock In \emph{Proceedings of the 3rd Workshop on Advances in Language and Vision Research (ALVR)}, pages 92--101, Bangkok, Thailand. Association for Computational Linguistics.

\bibitem[{Wei et~al.(2023)Wei, Wang, Schuurmans, Bosma, Ichter, Xia, Chi, Le, and Zhou}]{wei2023chainofthoughtpromptingelicitsreasoning}
Jason Wei, Xuezhi Wang, Dale Schuurmans, Maarten Bosma, Brian Ichter, Fei Xia, Ed~Chi, Quoc Le, and Denny Zhou. 2023.
\newblock \href {https://arxiv.org/abs/2201.11903} {Chain-of-thought prompting elicits reasoning in large language models}.
\newblock \emph{Preprint}, arXiv:2201.11903.

\bibitem[{Wieting and Kiela(2019)}]{wieting2019no}
John Wieting and Douwe Kiela. 2019.
\newblock No training required: Exploring random encoders for sentence classification.
\newblock \emph{arXiv preprint arXiv:1901.10444}.

\bibitem[{Xu et~al.(2024{\natexlab{a}})Xu, Tan, Wang, Yang, and Xu}]{xu-etal-2024-numcot}
Ancheng Xu, Minghuan Tan, Lei Wang, Min Yang, and Ruifeng Xu. 2024{\natexlab{a}}.
\newblock \href {https://doi.org/10.18653/v1/2024.findings-acl.848} {{NUMC}o{T}: Numerals and units of measurement in chain-of-thought reasoning using large language models}.
\newblock In \emph{Findings of the Association for Computational Linguistics: ACL 2024}, pages 14268--14290, Bangkok, Thailand. Association for Computational Linguistics.

\bibitem[{Xu et~al.(2024{\natexlab{b}})Xu, Fei, Pan, Liu, Lee, and Hsu}]{xu-etal-2024-faithful}
Jundong Xu, Hao Fei, Liangming Pan, Qian Liu, Mong-Li Lee, and Wynne Hsu. 2024{\natexlab{b}}.
\newblock \href {https://doi.org/10.18653/v1/2024.acl-long.720} {Faithful logical reasoning via symbolic chain-of-thought}.
\newblock In \emph{Proceedings of the 62nd Annual Meeting of the Association for Computational Linguistics (Volume 1: Long Papers)}, pages 13326--13365, Bangkok, Thailand. Association for Computational Linguistics.

\bibitem[{Zhang and Bowman(2018)}]{zhang2018language}
Kelly~W Zhang and Samuel~R Bowman. 2018.
\newblock Language modeling teaches you more syntax than translation does: Lessons learned through auxiliary task analysis.
\newblock \emph{arXiv preprint arXiv:1809.10040}.

\bibitem[{Zhao et~al.(2022)Zhao, Ziser, and Cohen}]{zhao-etal-2022-understanding}
Zheng Zhao, Yftah Ziser, and Shay Cohen. 2022.
\newblock \href {https://doi.org/10.18653/v1/2022.blackboxnlp-1.16} {Understanding domain learning in language models through subpopulation analysis}.
\newblock In \emph{Proceedings of the Fifth BlackboxNLP Workshop on Analyzing and Interpreting Neural Networks for NLP}, pages 192--209, Abu Dhabi, United Arab Emirates (Hybrid). Association for Computational Linguistics.

\bibitem[{Zhao et~al.(2024)Zhao, Ziser, and Cohen}]{zhao-etal-2024-layer}
Zheng Zhao, Yftah Ziser, and Shay~B Cohen. 2024.
\newblock \href {https://doi.org/10.18653/v1/2024.emnlp-main.847} {Layer by layer: Uncovering where multi-task learning happens in instruction-tuned large language models}.
\newblock In \emph{Proceedings of the 2024 Conference on Empirical Methods in Natural Language Processing}, pages 15195--15214, Miami, Florida, USA. Association for Computational Linguistics.

\bibitem[{Zhao et~al.(2023)Zhao, Ziser, Webber, and Cohen}]{zhao-etal-2023-joint}
Zheng Zhao, Yftah Ziser, Bonnie Webber, and Shay Cohen. 2023.
\newblock \href {https://doi.org/10.18653/v1/2023.findings-emnlp.851} {A joint matrix factorization analysis of multilingual representations}.
\newblock In \emph{Findings of the Association for Computational Linguistics: EMNLP 2023}, pages 12764--12783, Singapore. Association for Computational Linguistics.

\bibitem[{Zheng et~al.(2023)Zheng, Yang, Jiang, Lin, Lyu, She, and Wang}]{zheng-etal-2023-chain}
Mingyu Zheng, Hao Yang, Wenbin Jiang, Zheng Lin, Yajuan Lyu, Qiaoqiao She, and Weiping Wang. 2023.
\newblock \href {https://doi.org/10.18653/v1/2023.findings-emnlp.734} {Chain-of-thought reasoning in tabular language models}.
\newblock In \emph{Findings of the Association for Computational Linguistics: EMNLP 2023}, pages 11006--11019, Singapore. Association for Computational Linguistics.

\end{thebibliography}

\appendix

\section{Technical Details}
\label{app:technical-details}

\paragraph{Classification Models}
\begin{table}[b]
    \centering
    
    \begin{tabular}{ccc}

           \midrule
    \textbf{Cn-k12} &\textbf{AQuA} & \textbf{Olympiad} \\
   \midrule
          layer 31 & layer 32 & layer 13 \\
          layer 10 & layer 10  & layer 28\\
         layer 11 & layer 13 & layer 32\\
         layer 13& layer 12 & layer 10 \\
        layer 17& layer 5 & layer 3 \\
        \bottomrule
    \end{tabular}
    
\caption{Five most dissimilar layers using SVCCA (Singular Vector Canonical Correlation Analysis) between hidden representation with and with CoT prompt per layer, averaged over 1000 samples.}

\label{tab:top-5-similarity}
\end{table}
\label{app:classification-model}
We performed hyperparameter-tuning for each dataset (and each layer) when training the classification models. We illustrate the best hyperparameters in \autoref{tab:hyperparameter} and used them in our Prediction over time experiments.  

\begin{table*}[]
 \centering
 \begin{tabular}{lccccc}
     \toprule
    Dataset  & batch size & weight init  &  learning rate & optimizer & threshold\\
   \midrule
   \multicolumn{6}{c}{\textit{BERT (baseline)}} \\
   \midrule
   AQuA & 128 & {HE\_uniform} &{0.001} & \textbf{sgd}& {0.6} \\
   Olympiad & 128 &  {HE\_uniform} &{0.001}&{adam}& {0.6}\\
   Cn-k12 & 128 & -  &{0.001}&{adam} & 0.6 \\
   \bottomrule
    \multicolumn{6}{c}{\textit{Our Model}} \\
   \midrule
   AQuA & {32} &{HE\_uniform}  & {0.001} &\textbf{adam} & {0.5}\\
   Olympiad & 128 & {HE\_normal} &\textbf{0.001}&  {adam} & {0.5}\\
   Cn-k12 & {16} & {HE\_uniform} &{0.001}&{sgd}& {0.5} \\
   \bottomrule
 \end{tabular}
  \caption{Accuracy of the classification model before it starts generating and over time when model has generated x\% of the answer.}
   \label{tab:hyperparameter}
\end{table*}

\section{Additional Results}
\label{app:additional-results}
\begin{figure*}[]
    \centering
    \subfloat[\centering AQuA]{{\includegraphics[width=11cm]{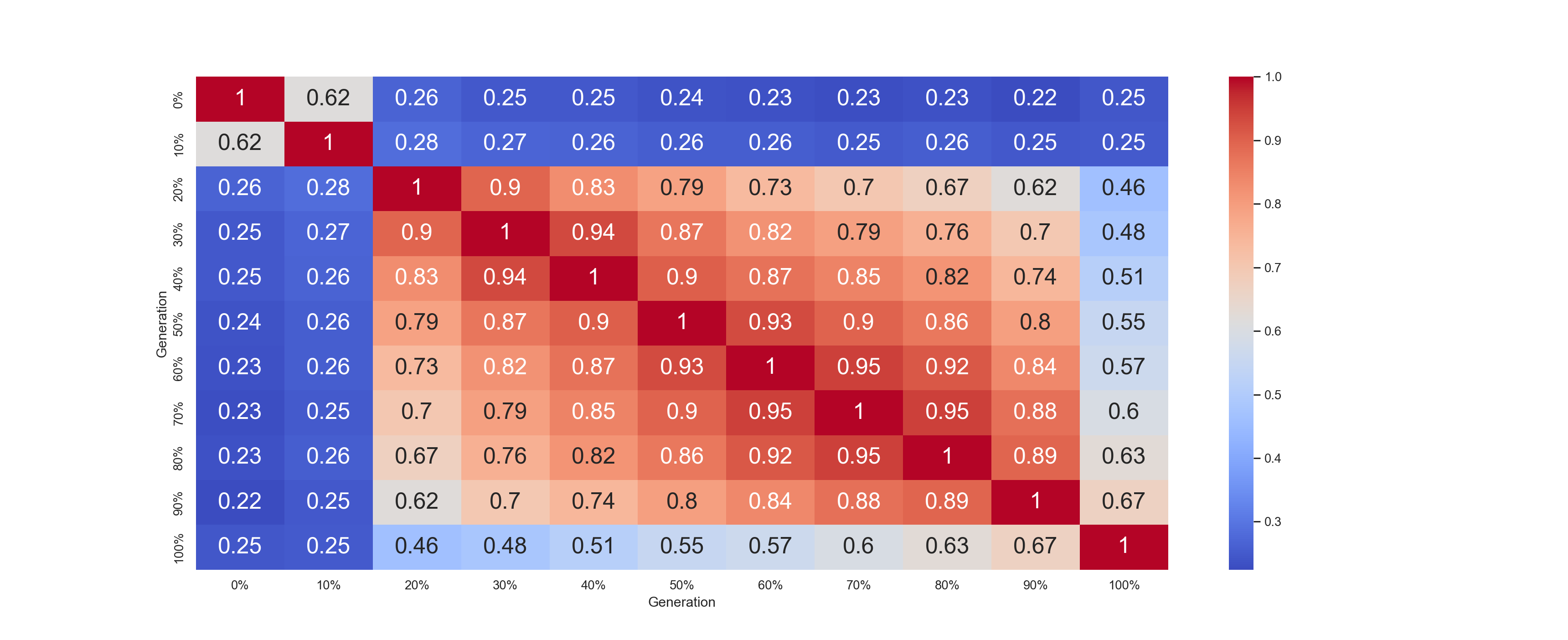} }} \\
    \subfloat[\centering Cn-k12]{{\includegraphics[width=11cm]{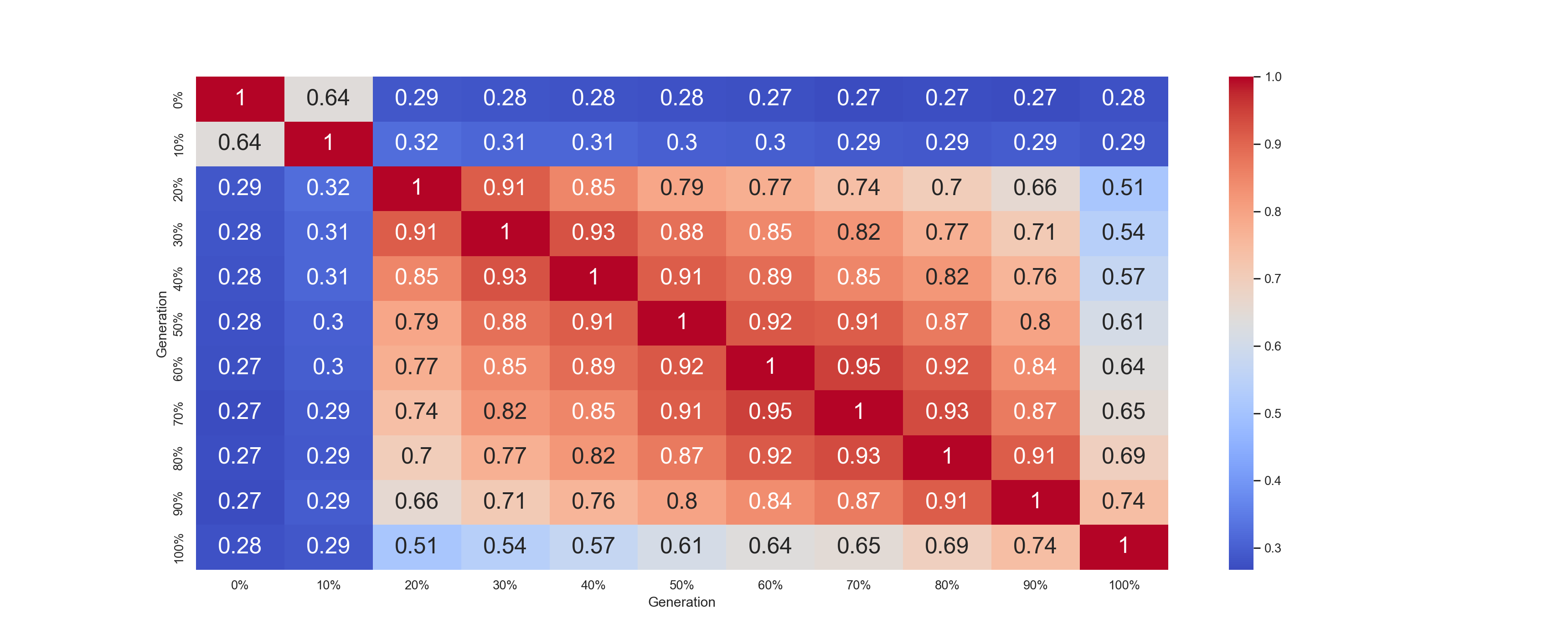} }} \\
    \subfloat[\centering Olympiad]{{\includegraphics[width=11cm]{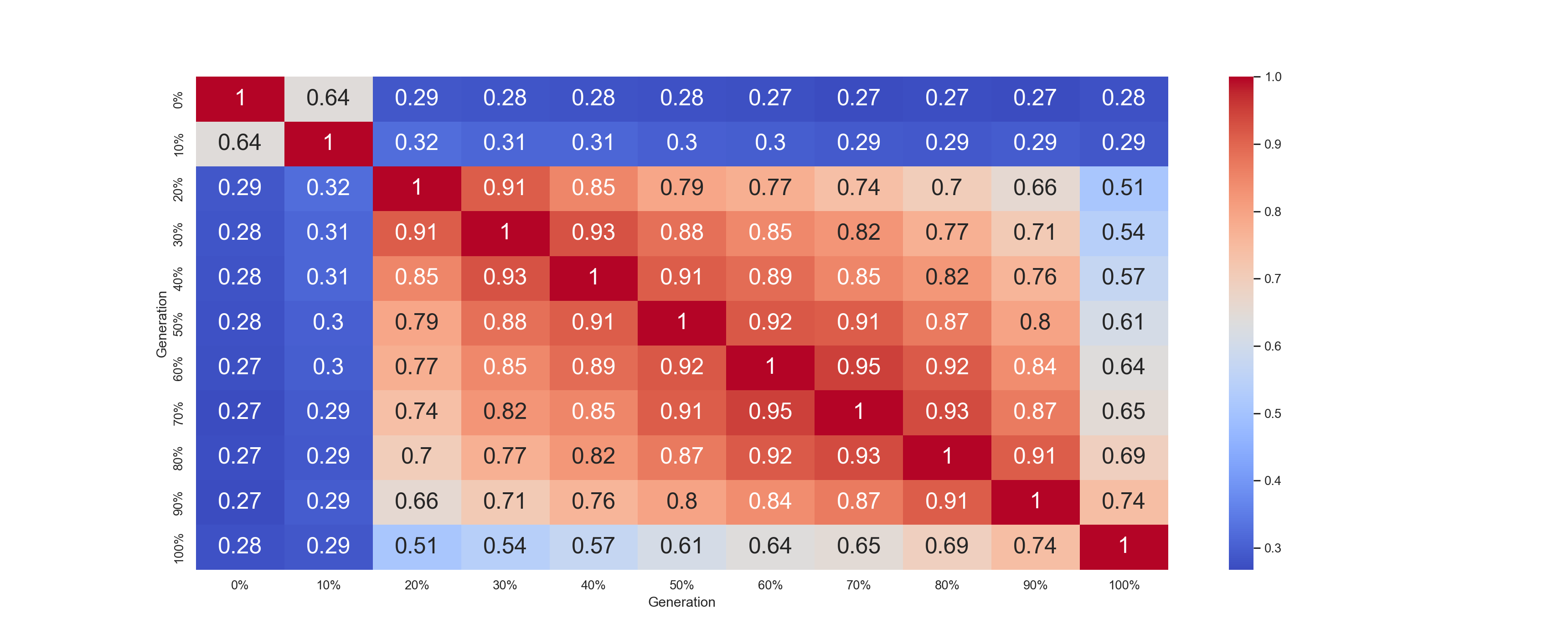} }}
    \caption{Similarity Scores using Singular Vector Canonical Correlation Analysis (SVCCA) of the internal representation of LLM through Prediction over Time using Layer 14. }
    \label{fig:pot_layer14_analysis}%
\end{figure*}

\paragraph{Complimentary Analysis for Probing}
We compute SVCCA scores between all possible combinations of representations reasoning step (10\%, 20\%, ..., 90\%) and with 100\% (CoT completion) and show them in \autoref{fig:pot_layer14_analysis}.

\paragraph{Early Stopping in CoT Reasoning}
\label{app:early-stopping}
We show an illustration of Early Stopping in CoT Reasoning and the LLM response in \autoref{tab:cot-interrupt}. It can be seen that despite being prompted to stop calculations to 30\% of it's original generatation, it still generates the correct answer.
\begin{table*}
\small
\centering
\begin{tabular}{p{15.4cm}l}
\toprule 
A high school has a total of 900 students, among which there are 300 freshmen, 200 sophomores, and 400 juniors. Now, a stratified sampling method is used to select a sample of 45 students. How many students should be selected from each grade, respectively?
\\
A: 15, 5, 25\\  
B: 15, 15, 15 \\ 
C: 10, 5, 30  \\
D: 15, 10, 20\\
Let's think step by step:\\
 To determine the number of students to be selected from each grade, we need to calculate the proportion of students in each grade and then apply this proportion to the total sample size of 45 students.\\

Step 1: Calculate the proportion of students in each grad\\ 
\\
\textit{Stop all computation and give me the correct answer in 2- 3 words, if you know it already.}\\
\\
Answer\: \textbf{15, 10, 20} \\
\midrule
Reference Answer\: The correct answer is D.\\
\midrule

\end{tabular}
\caption{An Illustration of a sample showing how we artificially halt LLM generation when it has generated 30\% of the answer.}
\label{tab:cot-interrupt}
\end{table*}

\paragraph{Classification Performance Breakdown}
\label{app:confusion-matrix}
Following our earlier evaluations, we perform an in-depth analysis of the classification model's performance on a test set of 1,000 examples, summarized in \autoref{tab:confusion_matrix}. Unlike the BERT baseline, which performs well on true positives (TP) only, our model excels at identifying both true negatives (TN) and true positives (TP). For example, on the Llama-3.1-8b generated Cn-k12 dataset, BERT identified 250 true negatives, while our model identified 317. Similarly, on AQuA, BERT's true negatives were 131, while our model achieved 177, and on Olympiad, BERT had 315 true negatives, compared to 388 for our method. These results show that our model not only performs well on true positives but also significantly outperforms BERT in detecting true negatives. This suggests that our method captures more relevant predictive signals, enabling a more nuanced understanding of the factors that drive Chain-of-Thought success.

\begin{table*}[h!]
    \centering

    \begin{tabular}{cccccc}
       \multicolumn{2}{c}{\textbf{Cn-k12}} & \multicolumn{2}{c}{\textbf{AQuA}} & \multicolumn{2}{c}{\textbf{Olympiad}} \\ 
         \toprule      
    \multicolumn{6}{c}{\textbf{\textit{\texttt{Llama-3.1-8B}}}} \\
        \toprule
   \multicolumn{6}{c}{\textit{BERT (baseline)}} \\
   \midrule
    \multicolumn{2}{c|}{250 (TN), 196 (FP)} & \multicolumn{2}{c|}{131 (TN), 306 (FP)} & \multicolumn{2}{c}{315 (TN), 159 (FP)} \\
    \multicolumn{2}{c|}{142 (FN), 412 (TP)} & \multicolumn{2}{c|}{159 (FN), 404 (TP)} & \multicolumn{2}{c}{135 (FN), 391 (TP)} \\
    \midrule
   \multicolumn{6}{c}{\textit{Our Model}} \\
      \midrule
    \multicolumn{2}{c|}{317 (TN), 129 (FP)} & \multicolumn{2}{c|}{177 (TN), 269 (FP)} & \multicolumn{2}{c}{388 (TN), 86 (FP)} \\
    \multicolumn{2}{c|}{180 (FN), 374 (TP)} & \multicolumn{2}{c|}{147 (FN), 416 (TP)} & \multicolumn{2}{c}{153 (FN), 373 (TP)} \\
        \toprule      
    \multicolumn{6}{c}{\textbf{\textit{\texttt{Mistral-7B}}}} \\
    \toprule
       \multicolumn{6}{c}{\textit{BERT (baseline)}} \\
   \midrule
    \multicolumn{2}{c|}{282 (TN), 221 (FP)} & \multicolumn{2}{c|}{314 (TN), 181 (FP)} & \multicolumn{2}{c}{318 (TN), 177 (FP)} \\
    \multicolumn{2}{c|}{124 (FN), 373 (TP)} & \multicolumn{2}{c|}{218 (FN), 287 (TP)} & \multicolumn{2}{c}{135 (FN), 370 (TP)} \\
    \midrule
   \multicolumn{6}{c}{\textit{Our Model}} \\
      \midrule
    \multicolumn{2}{c|}{328 (TN), 175 (FP)} & \multicolumn{2}{c|}{366 (TN), 129 (FP)} & \multicolumn{2}{c}{387 (TN), 108 (FP)} \\
    \multicolumn{2}{c|}{154 (FN), 343 (TP)} & \multicolumn{2}{c|}{244 (FN), 281 (TP)} & \multicolumn{2}{c}{178 (FN), 327 (TP)} \\
    \midrule
    \end{tabular}%
    \caption{Confusion matrix comparing BERT (baseline) and our model at $T=0$ on its respective balanced test set using the best-performing layers on the test set.}
    \label{tab:confusion_matrix}
\end{table*}
\paragraph{LLM Layer Analysis}
We also include the representation of LLM without any CoT reasoning in our experiments. In \autoref{tab:top-5-similarity}, we illustrate the 5 most dissimilar layers. Since they were both encoded using the same problems, the only different is the inclusion of CoT Reasoning, the disimilarity among the layers may just a more notable role in LLM's internal notion of CoT Reasoning.

\section{Dataset Examples}
\label{app:dataset-examples}
\paragraph{Success Prediction over Time}
An illustration of samples from the Success Prediction over Time method is shown in \autoref{tab:pot-sample} for 2 different generation steps.
\begin{table*}
\small
\centering
\begin{tabular}{p{15.4cm}l}
\textbf{T = 0 (Question + Prompt)}\\
\toprule 
Let the complex numbers $z_1$ and $z_2$ have corresponding points in the complex plane that are symmetric with respect to the real axis. Given that $z_1=1+i$, find $z_1z_2$. Let's think step by step: \\
\toprule

\textbf{T = 3 (Question + Prompt + 30\% of the Generation)}\\
\toprule 
Let the complex numbers $z_1$ and $z_2$ have corresponding points in the complex plane that are symmetric with respect to the real axis. Given that $z_1=1+i$, find $z_1z_2$.
Let's think step by step:
 To find $z_2$, we need to understand the symmetry of complex numbers with respect to the real axis. 

When a complex number is symmetric with respect to the real axis, its imaginary part is negated. 

Given $z_1 = 1 + i$, we can find $z_2$ by negating the imaginary part of $z_1$. 

S \\
\toprule

\textbf{T = 7 (Question + Prompt + 70\% of the Generation)}\\
\toprule 
Let the complex numbers $z_1$ and $z_2$ have corresponding points in the complex plane that are symmetric with respect to the real axis. Given that $z_1=1+i$, find $z_1z_2$.
Let's think step by step:
 To find $z_2$, we need to understand the symmetry of complex numbers with respect to the real axis. 

When a complex number is symmetric with respect to the real axis, its imaginary part is negated. 

Given $z_1 = 1 + i$, we can find $z_2$ by negating the imaginary part of $z_1$. 

So, $z_2 = 1 - i$.

Now, we need to find the product of $z_1$ and $z_2$. 

$z_1z_2 = (1 + i)(1 - i)$

Using the difference of squares formula $(a + b)(a - b) = a^2 - b^2$, we can simplify the expression:

$z_1z_2 = 1^2 - i^2$

Since $i^2 = -1$, we can substitute this value into the ex\\
\toprule
\end{tabular}
\caption{An Illustration of a sample from Cn-k12 dataset outlining how we construct the Prediction over Time dataset in our methodology.}
\label{tab:pot-sample}
\end{table*}



\end{document}